\documentclass[10pt,twocolumn,letterpaper]{article}

\usepackage{booktabs}
\usepackage{stfloats}

\usepackage[pagenumbers]{iccv} %

\newcommand{\greencheck}{\textcolor{green}{\checkmark}}
\newcommand{\redx}{\textcolor{red}{$\times$}}
\newcommand{\yellowtri}{\textcolor{yellow}{$\blacktriangle$}}
\definecolor{mylightblue}{rgb}{0.85, 0.90, 0.94}

\definecolor{C0}{rgb}{0.121569, 0.466667, 0.705882}
\definecolor{C1}{rgb}{1.000000, 0.498039, 0.054902}
\definecolor{C2}{rgb}{0.172549, 0.627451, 0.172549}
\definecolor{C3}{rgb}{0.839216, 0.152941, 0.156863}
\definecolor{C4}{rgb}{0.580392, 0.403922, 0.741176}
\definecolor{C5}{rgb}{0.549020, 0.337255, 0.294118}
\definecolor{C6}{rgb}{0.890196, 0.466667, 0.760784}
\definecolor{C7}{rgb}{0.498039, 0.498039, 0.498039}
\definecolor{C8}{rgb}{0.737255, 0.741176, 0.133333}
\definecolor{C9}{rgb}{0.090196, 0.745098, 0.811765}

\definecolor{iccvblue}{rgb}{0.21,0.49,0.74}
\usepackage[pagebackref,breaklinks,colorlinks,allcolors=iccvblue]{hyperref}

\def\paperID{12146} %
\def\confName{ICCV}
\def\confYear{2025}

\title{CapeLLM: Support-Free Category-Agnostic Pose Estimation with Multimodal Large Language Models}

\author{
  Junho Kim\textsuperscript{1}, \quad
  Hyungjin Chung\textsuperscript{1}\thanks{Corresponding authors}, \quad
  Byung-Hoon Kim\textsuperscript{1,2,*} \\
  \\
  \textsuperscript{1}EverEx \quad
  \textsuperscript{2}Yonsei University \\
  {\tt\small \{jh.kim, hj.chung, bh.kim\}@everex.co.kr}
}

\usepackage{multirow}
\usepackage{makecell}
\usepackage{colortbl}
\usepackage{longtable}

\begin{document}
\maketitle
\begin{abstract}
Category-agnostic pose estimation (CAPE) has traditionally relied on support images with annotated keypoints, a process that is often cumbersome and may fail to fully capture the necessary correspondences across diverse object categories. Recent efforts have explored the use of text queries, leveraging their enhanced stability and generalization capabilities. However, existing approaches often remain constrained by their reliance on support queries, their failure to fully utilize the rich priors embedded in pre-trained large language models, and the limitations imposed by their parametric distribution assumptions. To address these challenges, we introduce CapeLLM, the first multimodal large language model (MLLM) designed for CAPE. Our method only employs query image and detailed text descriptions as an input to estimate category-agnostic keypoints. Our method encompasses effective training strategies and carefully designed instructions for applying the MLLM to CAPE. Moreover, we propose an inference mechanism that further enhances the reasoning process for unseen keypoints. while flexibly modeling their underlying spatial distribution and uncertainty, allowing for adaptive refinement based on contextual cues. We conducted extensive experiments to apply the MLLM to CAPE effectively, focusing not only on the model architecture and prompt design but also on ensuring robustness across input variations. Our approach sets a new state-of-the-art on the MP-100 benchmark in the 1-shot and even 5-shot setting, marking a significant advancement in the field of category-agnostic pose estimation. Code is available \href{https://github.com/Junhojuno/CapeLLM}{here}.
\end{abstract}
\vspace{-0.3cm}

\section{Introduction}
\label{sec:intro}

Traditional pose estimation typically focuses on a single category, such as human~\cite{simplebaseline,hrnet,vitpose}, vehicle~\cite{carfusion,occlusionnet}, or animal~\cite{ap10k, macaquepose}. However, category-specific pose estimation lacks generalizability and cannot be applied to objects beyond the training set.

\begin{figure}[tb!]
  \centering
   \includegraphics[width=\linewidth, height=1.05\linewidth]{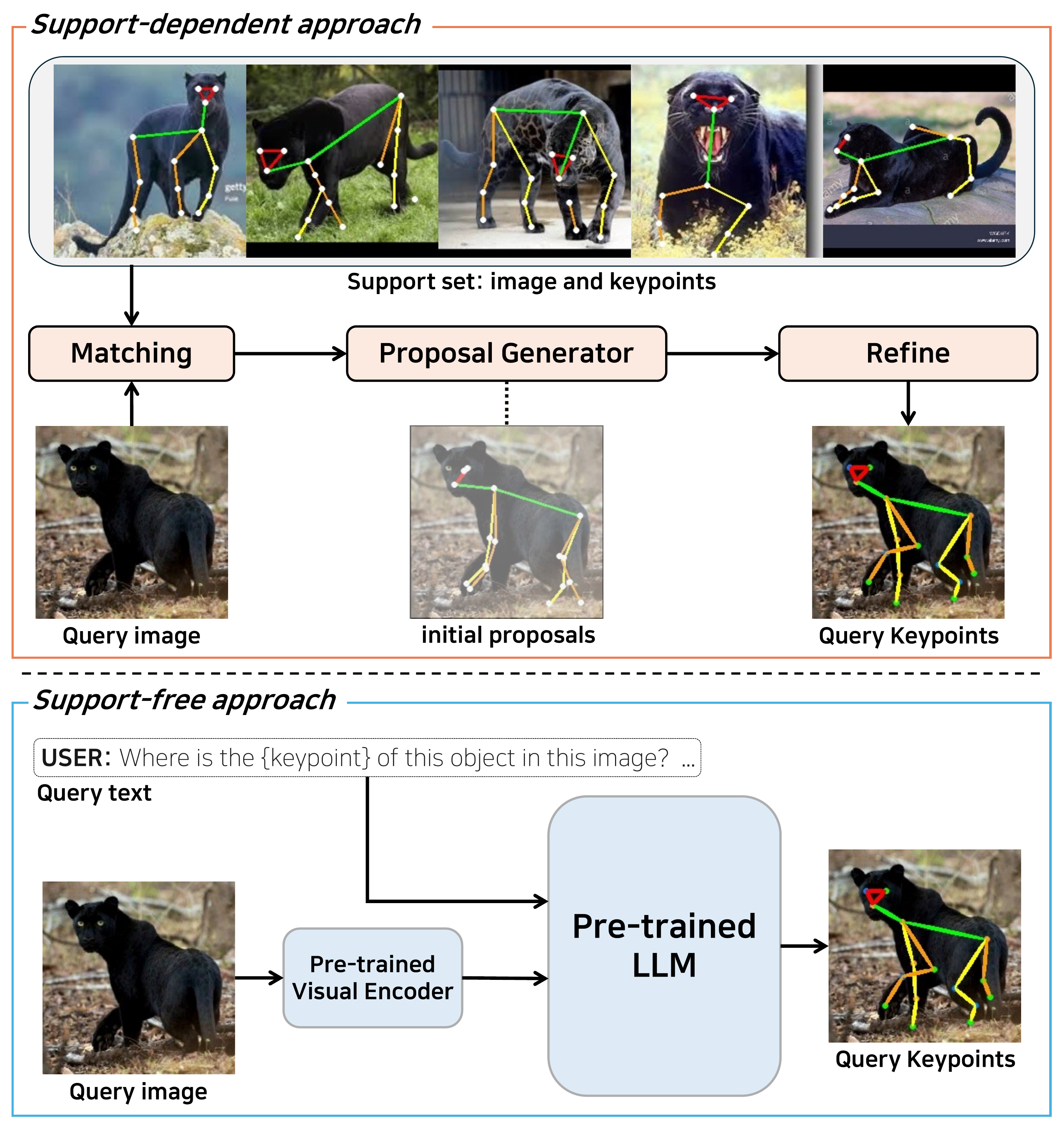}
   \caption{Architectural Comparison. Prior methods (top) are support-dependent approaches requiring support images and keypoint annotations, but ours (bottom) is support-free approach taking only text description of the keypoints in stead of the supports.}
   \vspace{-0.5cm}
   \label{fig:teaser}
\end{figure}

Category-Agnostic Pose Estimation (CAPE) methods address this issue~\cite{pomnet, capeformer, poseanything, capex, escape, meta-point, dynamic_support_info}. These methods predict keypoint positions of novel objects by employing the existing input image (called “query image”) with a set of supports. Typically, the support data comprises a pair: an image that belongs to the same category as the query but has a different pose from it, coupled with its corresponding keypoint annotations. 

Most of works in CAPE adopt a two-stage architecture~\cite{capeformer} that incorporates maximizing the similarity between query and support features and a process to refine the similarity to enhance performance consistency. Recently, the types of support data have been diversified. Skeletal structures, representing the connections between keypoints, are used as additional clues to figure out the keypoint location, achieving not only higher accuracy but also robustness against occlusions~\cite{poseanything}. Another type of support is to replace traditional support information with a textual one, such as a sequence of keypoint names~\cite{capex}, obtaining some freedom from the inherent reliance on the supports and therefore outperforming the conventional designs~\cite{pomnet,capeformer,poseanything}.

However, employing the support information comes with inherent drawbacks. Since this method aligns support and query images that differ in many aspects aside from belonging to the same category, inadequate generalization during training can cause the performance of the model to vary depending on the quality of the support data, even with the same query image. In addition to the inconsistency, it is often impractical to update the annotations whenever a new set of categories emerges.
To mitigate this issue, CapeX~\cite{capex} incorporated a text-based approach, where keypoint names are used instead of the support images. While this design choice enhanced the flexibility, CapeX is still heavily reliant on the skeletal representations. This poses a conundrum: Can text be the sole source of conditioning for CAPE, completely eliminating the need for additional auxiliary queries?

In this work, we answer this question with a positive by overcoming these challenges by introducing a Multimodal Large Language Model (MLLM) to CAPE. Our contributions can be summarized as follows:

\begin{itemize}
    \item We introduce CapeLLM, the first support-free framework in CAPE with advanced query-text comprehension capabilities, leveraging an MLLM. 
    \item We elucidate the design choices of using MLLMs for CAPE, from the design of the user query to specific training strategies. Interestingly, we reveal that tailored instructions are the key to unleashing the capabilities of MLLMs in CAPE.
    \item We propose dynamic round training, enabling spatial reasoning across multiple target poses.
    \item We propose a flexible decoding strategy that can implicitly model a general probability distribution over keypoints, rather than a fixed parameteric model such as a Gaussian
    \item We achieve state-of-the-art results on the MP-100 benchmark for CAPE, even outperforming the 5-shot accuracy of the previous art~\cite{poseanything}.
\end{itemize}

\section{Related Work}
\label{sec:related_work}

\subsection{Category-Agnostic Pose Estimation}

Approaches tackling CAPE can be broadly divided into two strategies: structural modifications~\cite{pomnet, capeformer, escape, meta-point, dynamic_support_info} and changes in the information provided~\cite{poseanything, capex}. The simplest structure was first proposed, which predicts keypoint positions by matching support data with the query image~\cite{pomnet}. This method connects different pieces of information using cross-attention and outputs similarity scores. While intuitive, it suffers from significant overfitting to the training data. To address this issue, an additional process was introduced to refine the matching results. This enhancement resolves the problems of the previous one-stage architecture and achieves higher performance by reducing dependence on the initial matching outcomes~\cite{capeformer}. 

Keeping the structure similar, performance can also be improved by increasing the amount of provided information. To consider interconnectivity among keypoints rather than treating them as independent entities, skeleton data is incorporated during training~\cite{poseanything}. By simply adding connection information, this approach not only achieved better quantitative performance than before but also demonstrated stable results in scenarios like occlusions. In addition to these methods, various approaches have been proposed, such as extracting and utilizing meaningful information from the provided support data~\cite{dynamic_support_info}, or defining meta-points~\cite{meta-point,escape} to link keypoints with similar semantics across multiple categories. However, these methods have not solved the fundamental problem that the models heavily rely on the auxiliary information, inducing discrepancy in prediction when changing the information, even with the same query image.

Recently, CapeX~\cite{capex} proposed to use text information for replacing the support query. Instead of using support images and their corresponding annotations, keypoint names for each category are defined as input and applied to the model. Trained by minimizing image-text similarity using a text encoder~\cite{gte}, CapeX achieves competitive performance. Although leveraging texts in CAPE was shown to be promising, CapeX still relies on the skeletal information proposed in a prior work~\cite{poseanything}. Without it, the performance of CapeX degrades heavily. Furthermore, due to the limited capability of the text encoder, only simple words were used as cues, without leveraging the rich semantic information embedded in language. On the other hand, as shown in Table~\ref{tab:difference_in_method}, CapeLLM effectively resolves both these issues by leveraging MLLMs, enabling the used of detailed text descriptions, while being completely free from other support queries.

\subsection{MLLMs for Visual Localization}

MLLMs have been widely incorporated into vision tasks, even in dense prediction tasks such as object detection and segmentation~\cite{pix2seq, visionllm, lenna, lisa,visionllmv2}, which were considered to be challenging for large language models (LLMs) to solve. Pix2Seq~\cite{pix2seq} introduces a novel perspective by transforming conventional object recognition problems into language modeling tasks. By eliminating overly task-specific modules, it streamlines the model architecture and enables language modeling by quantizing bounding boxes into input sequences. VisionLLM~\cite{visionllm} leverages LLMs to expand predefined vision-centric assignments into open-ended ones. \cite{lenna,lisa} use existing models as decoders instead of decoding text outputs. They make the outputs of LLMs be suitable for specific applications by employing foundation models in each vision domain~\cite{grounding_dino, sam}. 
A work that is closely related to ours is LocLLM~\cite{locllm}, which employs as input an image and a text instruction composed of the keypoint name and description. However, LocLLM is limited in that it is category-specific. As will be shown in later chapters, directly incorporating this scheme into CAPE offers several drawbacks\footnote{See, e.g.Tab.~\ref{tab:ablation/comparison_training_strategy}.}. To mitigate this issue, in the following, we investigate the optimal design choices for unleashing the capabilities of MLLM in CAPE.

\begin{table}[!t]
  \centering
  \resizebox{\columnwidth}{0.11\columnwidth}{
    \begin{tabular}{@{}lccc@{}}
      \toprule
      \textbf{Method} & \textbf{Category-Agnostic} & \textbf{Support-free} & \textbf{LLM-based}\\
      \midrule
      GraphCAPE~\cite{poseanything} & \greencheck & \redx & \redx \\
      CapeX~\cite{capex} & \greencheck & \yellowtri & \redx \\
      CapeLLM (\textbf{Ours}) & \greencheck & \greencheck & \greencheck\\
      \bottomrule
    \end{tabular}
  }
  \caption{Difference between prior methods and ours.}
  \label{tab:difference_in_method}
\end{table}

\section{CapeLLM}
\label{sec:method}

This section is structured as follows. In Sec.~\ref{sec:method/architecture}, we describe the model architecture for estimating novel keypoints from given textual descriptions. Then, we present how to design the keypoint names and their corresponding descriptions in Sec.~\ref{sec:method/instruction}. Finally, in Sec.~\ref{sec:method/training_strategy}, we introduce two distinct training strategies to endow the spatial reasoning capabilities to CapeLLM.

\begin{figure}[tb!]
  \centering
   \includegraphics[width=\linewidth, height=1.05\linewidth]{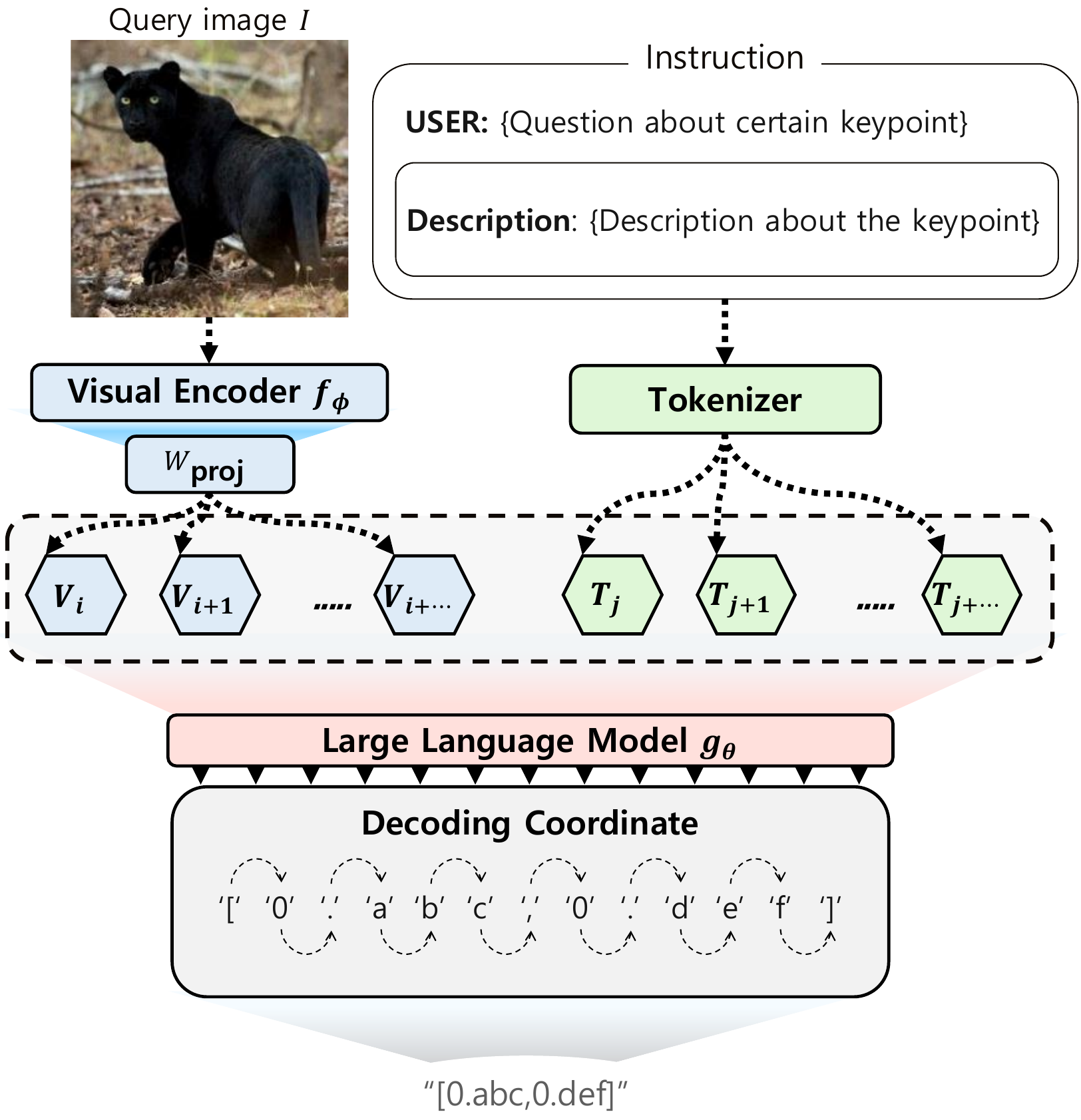}
   \caption{CapeLLM Architecture. CapeLLM consists of two pre-trained modules, visual encoder $f_{\phi}$ and LLM $g_{\theta}$. The visual tokens $V$ from visual encoder $f_{\phi}$ are fed into the LLM $g_{\theta}$ with the text tokens $T$. The decoding strategy predicts the keypoint coordinate directly as text generation.}
   \label{fig:model_arch}
\end{figure}

\subsection{CapeLLM Model Architecture}
\label{sec:method/architecture}

Similar to prior works~\cite{llava,locllm}, we utilize a pre-trained visual encoder in conjunction with a language model. To extract features across various categories, we adopt DINO-v2~\cite{dinov2}, which has been pre-trained on large-scale images with a variety of classes, as our visual encoder. For the language model, we select LLaMA3.1~\cite {llama3}, which has a powerful capability in a wide range of language tasks. The overall architecture is schematically illustrated in Figure~\ref{fig:model_arch}.

Our goal is to estimate keypoint coordinates of unseen categories only grounded on a query image and a text sequence that contains detailed information about the keypoint. To achieve this, the input image~$\mathbf{x} \in \mathbb{R}^{H \times W \times 3}$ is first divided into small patches, where $H$ and $W$ denote the height and width of the image, respectively. The patch images are processed through the visual encoder~$f_{\phi}$ to obtain the patch-processed image features $\Tilde{\mathbf{V}} \in \mathbb{R}^{N_{v} \times C}$, where $N_{v}$ is the number of patches and $C$ is the dimension of each patch. These patches are linearly transformed into image tokens~$\mathbf{V} \in \mathbb{R}^{N_{v} \times D}$ , via a simple learnable matrix~$\mathbf{W}_{\text{proj}} \in \mathbb{R}^{C \times D}$ in Eq~\eqref{eq:matmul}:

\begin{align}
    \Tilde{\mathbf{V}} &= f_{\phi}(\mathbf{x}) \label{eq:feature_extraction}\\
    \mathbf{V} &= \Tilde{\mathbf{V}} \mathbf{W}_{\text{proj}} \label{eq:matmul},
\end{align}

where $D$ represents the dimension of each image token.

These image tokens are prepended to the the query text token embeddings $\mathbf{T} \in \mathbb{R}^{N_{t} \times D}$, where $N_{t}$ denotes the number of text tokens, and then fed into the language model as the final input tokens $\mathbf{X} \in \mathbb{R}^{N \times D} $:

\begin{equation}
    \mathbf{X} = [\mathbf{V}; \mathbf{T}],
\end{equation}

where $;$ denotes concatenation of the two matrices along the token dimension, and $N = N_{v} + N_{t}$.

The decoder-only LLM~$g_{\theta}$ processes the input $\mathbf{X}$ to produce the output token matrix $\mathbf{Z}$ which has the same shape as $\mathbf{X}$:

\begin{equation}
    \mathbf{Z} = g_{\theta}(\mathbf{X}).
\end{equation}

Finally, following previous practices as in~\cite{llava,locllm}, a linear transformation with learnable parameters $\mathbf{W}_{\text{logit}} \in \mathbb{R}^{D \times M}$ outputs the final logits $\mathbf{Y} \in \mathbb{R}^{N \times M}$ for each token, where $M$ is the size of the vocabulary: 

\begin{equation}
  \mathbf{Y} = \mathbf{Z}\mathbf{W_{\text{logit}}}.
\end{equation}
While there are different strategies to decode the final predicted output of the keypoint, we choose a strategy that endows maximal flexibility in the output distribution. Specifically, we use a decodable floating-point representation by estimating with the following template: \texttt{[0.abc, 0.def]}, as shown in Fig.~\ref{fig:model_arch}. Here, each decimal point is represented by a separate token.

Concretely, let $y \in \mathbb{R}$ be either of the two scalar values that should be predicted. Then, CapeLLM approximates this value by factorizing it over $K = 3$ digit tokens, i.e.
\begin{align}
    p(y|\mathbf{x}) &\approx p_{\phi,\theta}(\mathbf{Y}_1, \mathbf{Y}_2, \dots, \mathbf{Y}_K|\mathbf{x}) \notag \\
    &= \prod_{k=1}^K p_{\phi,\theta}(\mathbf{Y}_k|\mathbf{x},\mathbf{Y}_1, \dots, \mathbf{Y}_{k-1}).
\label{eq:pose_posterior}
\end{align}
Then, one can show that (See Theorem 1 of \cite{song2025decoding}) CapeLLM models a truncated conditional density up to a resolution of $10^{-K}$. 
This leads to vastly enhanced flexibility, as opposed to the widely used Gaussian parametrization. See Sec.~\ref{sec:experiment/distribution_modeling} for detailed empirical results.

\subsection{Design of Instructions}
\label{sec:method/instruction}
Instructions exert a significant influence on MLLM's performance~\cite{llava, minigpt, instructblip, commit}. In CAPE, the categories and keypoints in the training set differ from those in the test set. Accordingly, most of the previous approaches~\cite{pomnet,capeformer,poseanything,meta-point,dynamic_support_info, escape} have utilized support sets and images with annotations, and recent work~\cite{capex} relies on keypoint names.  
We assume that using only keypoint names to infer their positions within images proves insufficient. To predict unseen keypoints, we opt for detailed descriptions of the keypoints and integrate this information into the instructions. Since some categories have densely defined keypoints, especially human faces, we manually design these descriptions to clearly delineate the differences between them. In crafting the descriptions, we avoid vague or ambiguous phrasing intentionally, instead providing precise details about the spatial positions and relationships among keypoints. For instance, when describing the ``front wheel” of a swivel chair, rather than saying ``starting from this wheel, the remaining wheels are located clockwise,” we expressed it as ``next to the fifth wheel and in front of the second wheel.”. We provide further details in Table~\ref{tab:appdx/def_name_desc/instruction_example}.

In the instructions, the answer is the keypoint coordinates, represented as normalized floating-point values. Although there are various coordinate representation methods, e.g., integer-valued binning~\cite{pix2seq}, or deviations from anchors~\cite{ranasinghe2024learning}, these approaches can be cumbersome due to the need for extra tokens or pre-defined anchors, while hampering flexibility. 
In contrast, we use the floating-point coordinate representation~\cite{shikra,locllm} as our solution, which not only is simple and robust without the need to incorporate additional tokens
but also leads to implicit modeling of arbitrary probability distributions of keypoints. See Sec.~\ref{sec:experiment/distribution_modeling} for details.
\vspace{-0.3cm}
\subsection{Training Strategy}
\label{sec:method/training_strategy}

\paragraph{Fixed round training}
To adapt the MLLM to the CAPE task, we introduce Fixed and Dynamic-Round Training strategies. Although the CAPE benchmark dataset, MP-100, covers a wide array of categories, each category contains only around 200 samples on average, which is considerably lower compared to other benchmarks (MSCOCO~\cite{mscoco}, MPII~\cite{mpii}, AP-10K~\cite{ap10k}). Therefore, the approach of matching an individual image with just a subset of keypoints, as used in LocLLM~\cite{locllm}, proves to be insufficient. Instead, we form (image, keypoints) pairs where every image is paired with all of its keypoints during training (``\textit{Fixed-Round}" strategy).
Initially, we partition the keypoints ($K_{\text{category}}$) into groups of $k$. Each set of keypoints is then combined with an image. It is important to note that we permit the repetition of images until every keypoint has been included in a pair. For example, in the bird category (where $K_{\text{bird}}$=15), if we group the keypoints in sets of 5 (i.e., $k$=5), a single bird image will produce 3 pairs. This approach guarantees that no keypoints are left unpaired during the training phase. Based on these individual pairs, we construct a multi-round conversation framework for model training. 
\vspace{-0.4cm}
\paragraph{Dynamic round training}
In addition to the Fixed-Round conversation method above, we also explore a ``\textit{Dynamic-Round}" strategy. Although both methods involve pairing an image with all the keypoints, the Dynamic-Round approach differs in that the number of keypoints linked to an image varies for each pair, unlike the fixed count of $k$ in the Fixed-Round method. This variability is intended to reinforce the reasoning process by utilizing information from other keypoints during prediction. The performance variations stemming from these two strategies are discussed in Section~\ref{sec:experiment/cumulative_reasoning}.
\vspace{-0.3cm}

\begin{figure*}[!t]
  \centering
   \includegraphics[width=0.85\linewidth]{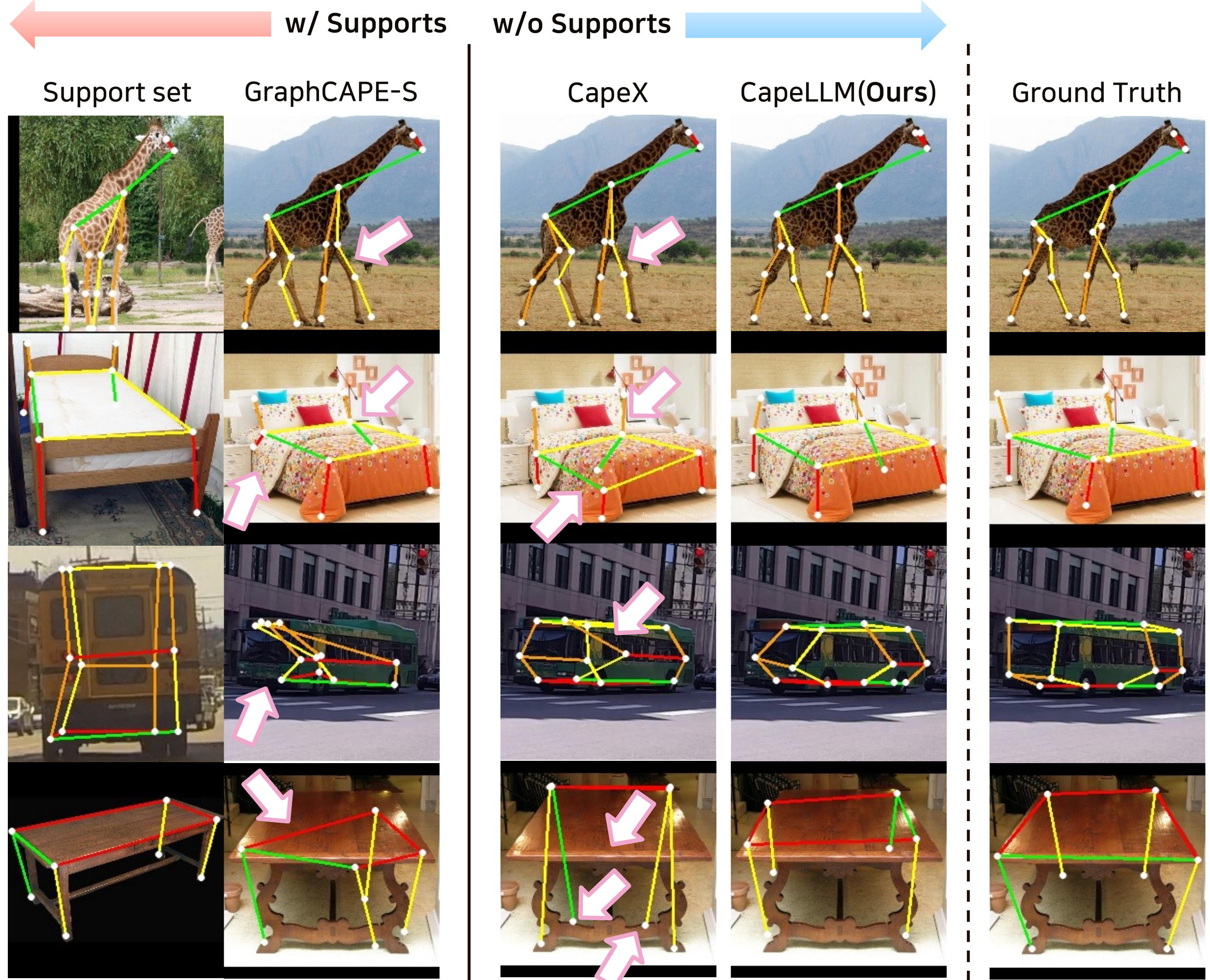}

   \caption{Qualitative results on MP-100. The support set is only used for GraphCAPE~\cite{poseanything}}
   \label{fig:qualitative_result}
\end{figure*}

\begin{table*}[!t]
  \centering
  \resizebox{0.9\textwidth}{!}{
    \begin{tabular}{@{}llcccccc@{}}
      \toprule
      \textbf{Evaluation dataset} & \textbf{Model} & \textbf{split1} & \textbf{split2} & \textbf{split3} & \textbf{split4} & \textbf{split5} & \textbf{Avg} \\
      \midrule
      \multirow{4}{*}{Support-Query Pairs} & GraphCAPE-S(1-shot)~\cite{poseanything} & 94.63 & 89.79 & 90.30 & 87.81 & 90.07 & 90.52 \\
      & GraphCAPE-S(5-shot)~\cite{poseanything} & 95.81 & 90.78 & \textbf{90.94} & 90.42 & 92.27 & 92.04 \\
      & CapeX~\cite{capex} & 95.29 & 91.08 & 88.94 & 89.83 & 92.96 & 91.62 \\
      \cmidrule(l){2-8}
      & \cellcolor{mylightblue} CapeLLM (\textbf{Ours}) & \cellcolor{mylightblue} \textbf{97.01} & \cellcolor{mylightblue} \textbf{92.40} & \cellcolor{mylightblue} 90.58 & \cellcolor{mylightblue} \textbf{90.90} & \cellcolor{mylightblue} \textbf{92.11} & \cellcolor{mylightblue} \textbf{92.60} \\
      \midrule
      \multirow{2}{*}{Only Query Images} & CapeX~\cite{capex} & 95.28 & 91.08 & 89.06 & 89.67 & \textbf{92.87} & 91.59 \\
      \cmidrule(l){2-8}
      & \cellcolor{mylightblue} CapeLLM (\textbf{Ours}) & \cellcolor{mylightblue} \textbf{96.98} & \cellcolor{mylightblue} \textbf{92.34} & \cellcolor{mylightblue}\textbf{90.57} & \cellcolor{mylightblue}\textbf{90.87} & \cellcolor{mylightblue}92.24 & \cellcolor{mylightblue}\textbf{92.60} \\
      \bottomrule
    \end{tabular}
  }
  \caption{PCK@0.2 on the MP-100 dataset. An method using support-set~\cite{poseanything} is usually evaluated on support-query pairs(``\textit{Support-Query Pairs}"), which means the the same query image can be assigned to different support images to form a pair. On the contrary, ``\textit{Only Query Images}" consists of unique queries, since it does not care about support-set.}
  \label{tab:quantitative_results_combined}
\end{table*}

\section{Experiment}
\label{sec:experiment}

\subsection{Environment}
\label{sec:experiment/implement_details}

\paragraph{Benchmark}
We utilize the MP-100 benchmark~\cite{pomnet}, aligning with prior methods~\cite{pomnet, capeformer, poseanything, capex, escape, meta-point, dynamic_support_info}. The benchmark comprises 100 categories and approximately 20,000 images. These categories are split into train, validation, and test sets in a 70/10/20 ratio without any category overlap. To ensure and identify robustness on unseen categories, the dataset is organized into five different splits, arranging each category to appear once in the test set. The number of keypoints per category varies significantly, ranging from 8 to 68. We discover that images in some categories (e.g., ``hand'') do not correctly correspond to keypoints; we rectify these discrepancies before proceeding with experiments. We use PCK@0.2 as a measurement of accuracy and compare both quantitative and qualitative results with two models as representatives: GraphCAPE~\cite{poseanything} and CapeX~\cite{capex}. GraphCAPE~\cite{poseanything} holds the highest 1-shot and 5-shot accuracy among models employing a query-support paired architecture(as seen in upper side of Figure~\ref{fig:teaser}). CapeX~\cite{capex}, based on the GraphCAPE framework, is a text-based query architecture utilizing keypoint names instead of support images and annotations. By connecting text features extracted via a text encoder\cite{gte} to image features, CapeX~\cite{capex} records higher 1-shot performance than GraphCAPE~\cite{poseanything}. Because of their overwhelmingly superior performance compared to other models, both quantitatively and qualitatively, we set these two methods as our baselines.

\paragraph{Implementation Details}
CapeLLM consists of a visual encoder, a projection layer, and an LLM. We choose DINO-v2~\cite{dinov2} as our vision encoder. Inspired by~\cite{llava}, the projection layer linking the vision and the text part is implemented as a linear layer. For the base language model, we use the pre-trained decoder-only model, LLaMA3.1~\cite{llama3}. We employ LoRA~\cite{lora} to fully leverage the capabilities of large-scale pre-trained models and adjust them to this task. Following the setting of~\cite{locllm}, we also apply LoRA~\cite{lora} to the query/value projection of the attention layers in both the visual encoder and the LLM, projecting them into the dimension of a rank 8. We selected AdamW as the optimizer and set the learning rate to $5 \times 10^{-4}$. The total number of training epochs was set to 12, with a warm-up phase over 3\% of the total training steps. Input images were resized to 224×224 pixels, and the default number of rounds included in each instruction was set to 4. We train our model on a Linux server equipped with four RTX-A6000 GPUs; instead of setting the batch size to 1 per GPU, we set the accumulation steps to 32.

\subsection{Main Results}
\label{sec:experiment/benchmark_results}

\paragraph{Quantitative Results}
While GraphCAPE~\cite{poseanything} configures the test dataset with support-query pairs, our method is completely support-free. For fair comparison, we make the same pairs as GraphCAPE~\cite{poseanything} and measure the performance only using the queries in the pairs(``Support-Query Pairs"). As shown in Table~\ref{tab:quantitative_results_combined}, our method outperforms the 1-shot accuracy of GraphCAPE~\cite{poseanything} over 1\%p and further achieves 0.56\%p higher accuracy than its 5-shot performance. Note that the information provided for 5-shot CAPE is strictly larger than in our case. Remarkably, this result demonstrates that even without any support, it is possible to attain superior results in CAPE.
We also compare our model with CapeX~\cite{capex} using about 2,000 query images from each split(``Only Query Images"). In Table~\ref{tab:quantitative_results_combined}. we achieve approximately 1\%p higher accuracy than CapeX~\cite{capex}, achieving state-of-the-art.

\begin{figure}[!t]
  \centering
   \includegraphics[width=\linewidth, keepaspectratio]{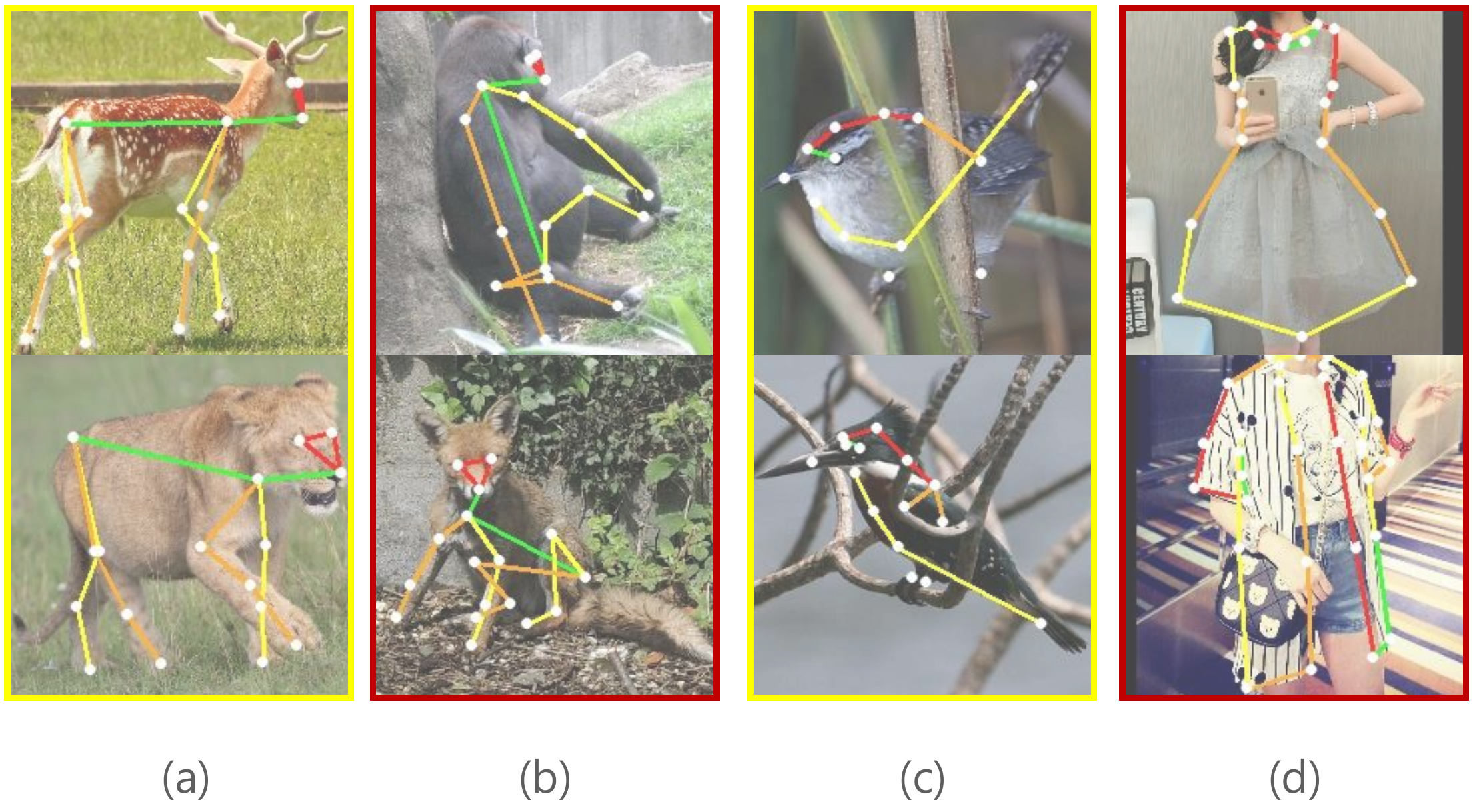}
   \caption{Qualitative results in diverse occlusions. There are diverse cases of occlusion: (a) Self-occlusion, (b) self-occlusion with articulated pose, (c) occlusion-by-object, and (d) self-and-object occlusion. CapeLLM performs robustly even in such challenging cases with occlusion.}
   \vspace{-0.5cm}
   \label{fig:qualitative_results_occlusion}
\end{figure}

\paragraph{Qualitative Results}
As can be seen in Figure~\ref{fig:qualitative_result}, we demonstrate that our CapeLLM is superior to conventional methods~\cite{poseanything, capex} across various categories. Specifically, in \textit{animal body}, remarkable improvements have been found in end joints, such as knees and paws, which the preceding approaches~\cite{poseanything, capex} have struggled to predict. In other categories, prior methods~\cite{poseanything, capex} show imprecise results. For instance, GraphCAPE~\cite{poseanything} fails due to differences of pose in the supports, and CapeX~\cite{capex} confuses the front and rear joints. Our method, however, not only produces more accurate results than previous approaches~\cite{poseanything, capex}, but also works as intended based on the given instructions. 
In Figure~\ref{fig:qualitative_results_occlusion}, we further show several challenging cases where there exists sufficient occlusion in the input image. Remarkably, CapeLLM performs robustly even under these challenging scenarios, without the use of skeletal structure~\cite{poseanything}, which are often used to explicitly enhance the robustness under occlusion.

\begin{table}[htbp!]
    \resizebox{\linewidth}{!}{
        \begin{tabular}{c|c|cc|ccc}
            \toprule
            \multirow{2}{*}{\textbf{Method}} & \multirow{2}{*}{\textbf{Training}} & \multicolumn{2}{c}{\textbf{Inference}} & \multicolumn{3}{c}{\textbf{Metric}} \\
            & & Single & Cumulative & PCK\@0.15 & PCK\@0.20 & PCK\@0.25\\
            \cmidrule(lr){1-2}
            \cmidrule(lr){3-4}
            \cmidrule(lr){5-7}
            \multirow{2}{*}{LocLLM-style~\cite{locllm}} & \multirow{2}{*}{-} & \greencheck & \redx & 91.00 & 94.85 & 96.98 \\
            & & \redx & \greencheck & 90.56 & 93.39 & 95.03 \\
            \midrule
            \multirow{4}{*}{Ours} & \multirow{2}{*}{Fixed} & \greencheck & \redx & \cellcolor{mylightblue} 95.26 & \cellcolor{mylightblue} 96.98 & \cellcolor{mylightblue} 97.90 \\
            & & \redx & \greencheck & 93.02 & 94.59 & 95.42 \\
            \cmidrule(l){2-7}
            & \multirow{2}{*}{Dynamic} & \greencheck & \redx & 94.31 & 96.05 & 97.26 \\
            & & \redx & \greencheck & \textbf{95.62} & \textbf{97.28} & \textbf{98.27} \\
            \bottomrule
        \end{tabular}
    }
    \caption{Result of cumulative reasoning. \colorbox{mylightblue}{Default config}.}
    \label{tab:ablation/cumulative_reasoning}
\end{table}

\subsection{Cumulative Reasoning for Pose Estimation}
\label{sec:experiment/cumulative_reasoning}
Inspired by the previous works related to Chain-of-Thought (CoT)~\cite{cot,llava,llava-cot}, we investigate whether the information from certain keypoints influences the estimation of others. Specifically, we devise an inference mechanism, called ``\textit{Cumulative Reasoning}", for helping to predict keypoint coordinates more precisely using the implicit capability of MLLMs, and analyze its effectiveness comparing with default process, single-round inference (``Single” in Table~\ref{tab:ablation/cumulative_reasoning}). 
For this reasoning task, the prompt employed for predicting each keypoint is prepended to the prompt for the subsequent keypoints, thereby establishing a cumulative context. As detailed in Table~\ref{tab:ablation/cumulative_reasoning}, while the LocLLM~\cite{locllm}-style and the fixed-training scheme result in decreased performance (see the 2nd and 4th rows), dynamic-round training strategy not only encourages the model to achieve better results than single-round inference by 1.2\%p (see the last row), but also outperforms the fixed-round strategy in Table~\ref{tab:ablation/comparison_training_strategy}. This experiment suggests that CapeLLM can extract richer spatial and relational cues based on the accumulated information from other keypoints, which in turn enables it to reason about the target keypoint’s position.

\begin{figure}[tb!]
  \centering
   \includegraphics[width=\linewidth, keepaspectratio]{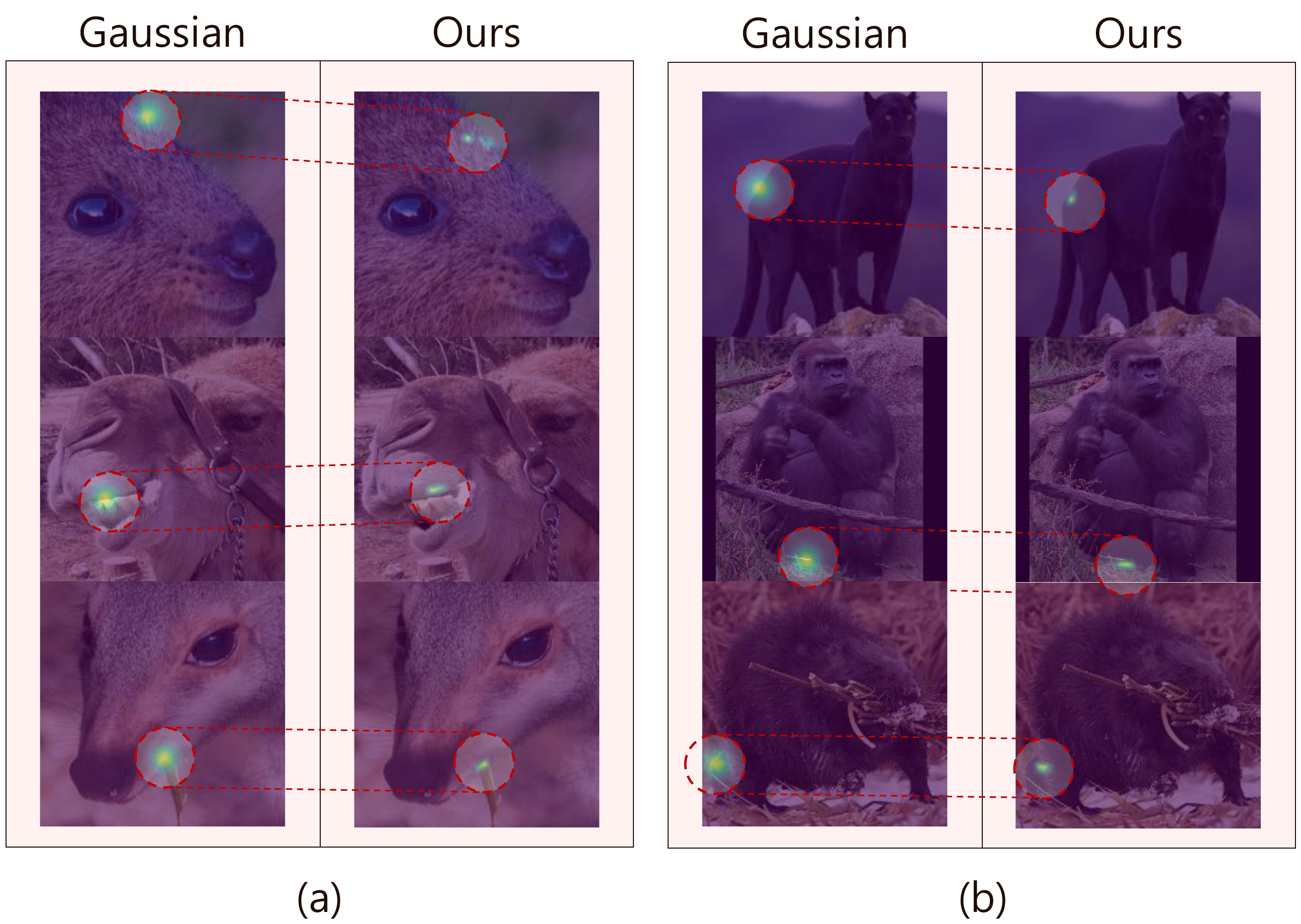}
   \caption{Distribution modeling for keypoints. 
   ``Gaussian'' displays a fixed-variance Gaussian around the ground-truth target point. ``Ours'' displays the distribution modeled by CapeLLM, achieved by sampling multiple points, and using kernel density estimation.
   In (a) \textbf{animal face}, (top) \textit{top left side of the left eye}; (mid) \textit{right side of the lip}; (bottom) \textit{left side of the lip}. In (b) \textbf{animal body}, (top) \textit{root of tail}; (mid) \textit{right back paw}; (bottom) \textit{left knee}.}
   \label{fig:distribution_modeling}
\end{figure}

\subsection{Implicit Distribution Modeling}
\label{sec:experiment/distribution_modeling}

Many previous CAPE methods are based on heatmap-based learning, and most of them rely on parametric distribution modeling, among them predominantly a Gaussian distribution. It is also often the case that a fixed variance is used~\cite{luo2021rethinking}, further constraining the situation. 
While such a strategy is known to be robust than point estimates, it also greatly limits the capacity of the neural network by constraining the output distribution. This strategy is also often physically implausible, as the points to be estimated are often located near the edge of the support. While convolving the point with a Gaussian would induce some probabilities in all the nearby regions, it should ideally allocate the probability of 0 for out-of-bounds. (See Fig.~\ref{fig:distribution_modeling} for several examples of this). On the other hand, CapeLLM generates the output coordinates
in an unconstrained fashion, which can approximate the true posterior with arbitrary precision~\cite{song2025decoding}. Using this property, we examine a density estimation for keypoints based on our decoding process, (``decoding-based method") and compare this method with the prevalent Gaussian modeling.
As illustrated in Figure~\ref{fig:distribution_modeling}, leveraging the inherent randomness in the MLLM's decoding strategy seems to make the points located closer to the ground truth point. Also, while the distribution of Gaussian modeling stretches across the background, that of the decoding-based method are kept within the foreground. This observation implies that the decoding-based modeling can be another option to build a density function of keypoint and resolve the inherent constraints in the conventional method, such as fixed-variance Gaussian~\cite{luo2021rethinking}.

\begin{table}[htbp!]
  \centering
  \resizebox{\linewidth}{!}{
    \begin{tabular}{c|c|ccc}
      \toprule
      \textbf{Method} & \textbf{Training} & \textbf{PCK@0.05} & \textbf{PCK@0.2} & \textbf{mPCK} \\
      \midrule
      LocLLM-style~\cite{locllm} & - & 55.15 & 94.85 & 84.00 \\
      \midrule
      \multirow{2}{*}{Ours} & \cellcolor{mylightblue} Fixed & \cellcolor{mylightblue} \textbf{78.43} & \cellcolor{mylightblue} \textbf{96.98} & \cellcolor{mylightblue} \textbf{91.98} \\
      \cmidrule(l){2-5}
      & Dynamic & 76.55 & 96.05 & 90.92 \\
      \bottomrule
    \end{tabular}
  }
  \vspace{-0.3cm}
  \caption{Result in different training strategies. \colorbox{mylightblue}{Default config}.}
  \label{tab:ablation/comparison_training_strategy}
  \vspace{-0.3cm}
\end{table}

\subsection{Ablation Studies}
\label{sec:experiment/ablation}

As CapeLLM is, to the best of our knowledge, the first work that uses MLLM for CAPE, in this section, we conduct extensive ablation studies to reveal the crucial components that lead to the superior performance of our method. We further show that CapeLLM is robust to variations in the input image resolution and the decoding strategy. Further ablations are deferred to the appendix.

\paragraph{Training strategy}
To investigate the effectiveness of our training schemes, we compare them with the training strategy from LocLLM~\cite{locllm}. The key difference between these strategies lies in the process to construct image–keypoint pairs. Our findings reveal that our approaches outperform the approach in LocLLM across all metrics. Notably, the fixed-round training demonstrates an approximately 8\%p improvement in mPCK, indicating that explicitly connecting all keypoints with a single image is more suitable for leveraging MLLMs in CAPE. Here, we determine $k$ value in fixed-round training empirically; the corresponding results are provided in the Appendix.

\begin{table}[htbp!]
  \centering
    \resizebox{0.9\linewidth}{!}{
        \begin{tabular}{ccccccc}
            \toprule
            \textbf{Description Type} & \textbf{PCK@0.05} & \textbf{PCK@0.2} & \textbf{mPCK} \\
            \midrule
            Vague & 69.82 & 91.97 & 85.98 \\
            \midrule
            \rowcolor{mylightblue} Spatial \& Relational & \textbf{78.43} & \textbf{96.98} & \textbf{91.98} \\
            \bottomrule
            \end{tabular}
    }
    \caption{Comparison with vague descriptions. \colorbox{mylightblue}{Our method}.}
    \vspace{-0.3cm}
    \label{tab:ablation/vague_description}
\end{table}

\begin{table}[htbp!]
    \centering
    \resizebox{0.9\linewidth}{!}{
        \begin{tabular}{ccccc}
            \toprule
            \textbf{Name} & \textbf{Description} & \textbf{PCK\@0.05} & \textbf{PCK\@0.20} & \textbf{PCK\@0.25} \\
            \midrule
            \rowcolor{mylightblue} \redx & \redx & \textbf{78.43} & \textbf{96.98} & \textbf{97.90} \\
            \greencheck & \redx & 78.24 & 96.94 & 97.83 \\
            \redx & \greencheck & 70.11 & 96.67 & 97.69 \\
            \greencheck & \greencheck & 66.16 & 96.09 & 97.26 \\
            \bottomrule
            \end{tabular}
    }
    \vspace{-0.3cm}
    \caption{Robustness to variation in input. \colorbox{mylightblue}{Default config}.}
    \label{tab:ablation/change_description_style}
\end{table}

\begin{table}[htbp!]
    \resizebox{\linewidth}{!}{
        \begin{tabular}{c|ccc|cc}
            \toprule
            \multicolumn{1}{c}{\textbf{In Training}} & \multicolumn{3}{c}{\textbf{In Inference}} & \multicolumn{2}{c}{\textbf{Metric}} \\
            Random Replaced & Detail Desc & Replaced & Removed & PCK\@0.20 & mPCK\\
            \cmidrule(lr){1-1}
            \cmidrule(lr){2-4}
            \cmidrule(lr){5-6}
            \rowcolor{mylightblue} \redx & \greencheck & \redx & \redx & \textbf{96.98} & \textbf{91.98} \\
            \redx & \redx & \greencheck & \redx & 95.28 & 86.66 \\
            \redx & \redx & \redx & \greencheck & 95.15 & 85.69 \\ 
            \midrule
            \greencheck & \greencheck & \redx & \redx & 96.65 & 89.87 \\
            \greencheck & \redx & \greencheck & \redx & 96.39 & 89.65 \\
            \greencheck & \redx & \redx & \greencheck & 96.07 & 88.84 \\
            \bottomrule
            \end{tabular}
    }
    \vspace{-0.3cm}
    \caption{Performance with varying descriptions. \colorbox{mylightblue}{Default config}.}
    \label{tab:ablation/drop_description}
    \vspace{-0.5cm}
\end{table}

\paragraph{Choice of instructions}
To investigate the impact of instructions on the performance, we define two types of variations in the instruction: (1) change in the keypoint names and descriptions, (2) the omission of keypoint description. 
Initially, we consider a scenario in which vague keypoint descriptions are provided, as mentioned in Sec.~\ref{sec:method/instruction}, and convert descriptions with the ambiguous ones in training. We find that incorporating detailed spatial and relational information among keypoints yields an improvement of up to 6\%p in mPCK compared to the baseline (Table~\ref{tab:ablation/vague_description}). This finding highlights the critical role of learning spatial positioning and contextual relationships for accurately predicting keypoint coordinates.
We also assess the model’s stability when confronted with input styles that differ from those seen during training. By utilizing GPT-4o, we prompt it to convert the keypoint names and descriptions in the test set into a simplified format that preserves their original meaning. According to Table~\ref{tab:ablation/change_description_style}, CapeLLM exhibits notably consistent performance across these variations.
Further, we scrutinize two cases where the descriptions are excluded: (1) employing a simple QA method to query keypoint coordinates without any description, (2) where the description is replaced with the statement “There is no description to refer to.”. In all cases, CapeLLM maintains relatively stable performance, and its robustness can be further enhanced when taking advantage of the random substitution of descriptions during training (Table~\ref{tab:ablation/drop_description}).

\begin{table}[htbp!]
    \centering
    \resizebox{0.8\linewidth}{!}{
        \begin{tabular}{ccccccc}
            \toprule
            \textbf{Resolution} & \textbf{PCK\@0.05} & \textbf{PCK\@0.20} & \textbf{PCK\@0.25} \\
            \midrule
            \rowcolor{mylightblue} 224$\times$224 & 78.43 & 96.98 & 97.90 \\
            238$\times$238 & \textbf{78.68} & 97.07 & 97.93 \\
            252$\times$252 & 78.45 & \textbf{97.11} & \textbf{97.99} \\
            336$\times$336 & 78.25 & 96.85 & 97.81 \\
            \bottomrule
            \end{tabular}
    }
    \vspace{-0.3cm}
    \caption{Robustness to larger image resolution. \colorbox{mylightblue}{Default config}.}
    \label{tab:ablation/larger_image_resolutions}
    \vspace{-0.5cm}
\end{table}

\paragraph{Larger image resolution}
We investigate how the performance varies when varying the input image size to be larger than the default 224$\times$224 resolution. As presented in Table~\ref{tab:ablation/larger_image_resolutions}, across many different input image sizes up to 336$\times$336, CapeLLM retains its robust performance.

\begin{table}[htbp!]
    \centering
    \resizebox{\linewidth}{!}{ %
        \begin{tabular}{c|cccccc}
            \toprule
            \textbf{Decoding Strategy} & \textbf{PCK@0.2} & \textbf{PCK@0.25} & \textbf{mPCK} \\
            \midrule
            \rowcolor{mylightblue} Greedy Search & 0.9698 & 0.9790 & 0.9198 \\
            \midrule
            Sampling & 0.9606 & 0.9726 & 0.8958 \\
            \midrule
            Sampling (\texttt{t=0.6}) & 0.9669 & 0.9776 & 0.9095 \\
            \midrule
            Top-3 Sampling & \textbf{0.9707} & \textbf{0.9792} & \textbf{0.9200} \\
            \midrule
            Top-5 Sampling & 0.9705 & 0.9791 & 0.9197 \\
            \midrule
            Nucleus Sampling (\texttt{p=0.92}) & 0.9669 & 0.9777 & 0.9071 \\
            \midrule
            Contrastive Search (\texttt{a=0.6}) & 0.9698 & 0.9790 & 0.9198 \\
            \bottomrule
        \end{tabular}
    }
    \vspace{-0.3cm}
    \caption{Performance comparison of different decoding strategies. \colorbox{mylightblue}{Default config}.}
    \label{tab:ablation/decoding_strategies}
    \vspace{-0.3cm}
\end{table}
\vspace{-2mm}
\paragraph{Decoding strategy}
As our decoding strategy is directly based on LLM token decoding, we can leverage many different standard LLM sampling strategies to sample from the posterior $p(y|\mathbf{x})$ in \eqref{eq:pose_posterior}. While we resort to greedy decoding for all of our main experiments for simplicity, we explore different strategies in Tab.~\ref{tab:ablation/decoding_strategies}. Here, we observe that CapeLLM is again robust to different sampling strategies, achieving state-of-the-art performance in {\em all} cases. We further find that with Top-$k$ sampling, we are able to achieve even better performance. This suggests that exploring more advanced decoding strategies could further enhance CapeLLM, which we leave as an avenue for future work.

\vspace{-0.2cm}
\section{Conclusion}

We introduce CapeLLM, the first fully support-free MLLM-based method for CAPE. By leveraging the reasoning capabilities of a pre-trained LLM, CapeLLM achieves state-of-the-art performance without requiring any support images or annotations, fundamentally challenging the paradigm of prior CAPE approaches.
Our method integrates MLLMs with a lightweight projection layer, allowing seamless alignment between visual and textual modalities. To fully harness MLLMs for CAPE, we design structured keypoint instructions, providing explicit spatial and relational descriptions that enable robust keypoint estimation across unseen categories. Furthermore, we introduce two novel training strategies—fixed-round and dynamic-round training—which not only improve category-agnostic keypoint prediction but also enhance cumulative reasoning, allowing the model to refine its predictions iteratively based on contextual information.
We believe CapeLLM serves as a foundational step toward the broader application of MLLMs in spatial reasoning and structured perception tasks. Future work can explore more advanced decoding strategies, multi-modal extensions, and scalability to real-world scenarios, paving the way for next-generation keypoint estimation models driven by MLLMs.

{
    \small
    \bibliographystyle{ieeenat_fullname}
    \bibliography{main}

\begin{thebibliography}{43}
\providecommand{\natexlab}[1]{#1}
\providecommand{\url}[1]{\texttt{#1}}
\expandafter\ifx\csname urlstyle\endcsname\relax
  \providecommand{\doi}[1]{doi: #1}\else
  \providecommand{\doi}{doi: \begingroup \urlstyle{rm}\Url}\fi

\bibitem[AI(2024)]{Llama3.2}
Meta AI.
\newblock {Llama 3.2: Vision Edge for Mobile Devices}, 2024.
\newblock \url{https://ai.meta.com/blog/llama-3-2-connect-2024-vision-edge-mobile-devices/}.

\bibitem[Andriluka et~al.(2014)Andriluka, Pishchulin, Gehler, and Schiele]{mpii}
Mykhaylo Andriluka, Leonid Pishchulin, Peter Gehler, and Bernt Schiele.
\newblock {2D Human Pose Estimation: New Benchmark and State of the Art Analysis}.
\newblock In \emph{Proceedings of the IEEE Conference on Computer Vision and Pattern Recognition (CVPR)}, 2014.

\bibitem[Chen et~al.(2024)Chen, Yan, Fang, and Niu]{meta-point}
Junjie Chen, Jiebin Yan, Yuming Fang, and Li Niu.
\newblock {Meta-Point Learning and Refining for Category-Agnostic Pose Estimation}.
\newblock In \emph{Proceedings of the IEEE/CVF Conference on Computer Vision and Pattern Recognition}, pages 23534--23543, 2024.

\bibitem[Chen et~al.(2023)Chen, Zhang, Zeng, Zhang, Zhu, and Zhao]{shikra}
Keqin Chen, Zhao Zhang, Weili Zeng, Richong Zhang, Feng Zhu, and Rui Zhao.
\newblock {Shikra: Unleashing Multimodal LLM's Referential Dialogue Magic}.
\newblock \emph{arXiv preprint arXiv:2306.15195}, 2023.

\bibitem[Chen et~al.(2022)Chen, Saxena, Li, Fleet, and Hinton]{pix2seq}
Ting Chen, Saurabh Saxena, Lala Li, David~J. Fleet, and Geoffrey Hinton.
\newblock {Pix2seq: A Language Modeling Framework for Object Detection}.
\newblock In \emph{International Conference on Learning Representations}, 2022.

\bibitem[Dai et~al.(2023)Dai, Li, Li, Tiong, Zhao, Wang, Li, Fung, and Hoi]{instructblip}
Wenliang Dai, Junnan Li, Dongxu Li, Anthony Tiong, Junqi Zhao, Weisheng Wang, Boyang Li, Pascale Fung, and Steven Hoi.
\newblock {InstructBLIP: Towards General-purpose Vision-Language Models with Instruction Tuning}.
\newblock In \emph{Thirty-seventh Conference on Neural Information Processing Systems}, 2023.

\bibitem[Darcet et~al.(2024)Darcet, Oquab, Mairal, and Bojanowski]{dinov2-reg}
Timoth{\'e}e Darcet, Maxime Oquab, Julien Mairal, and Piotr Bojanowski.
\newblock {Vision Transformers Need Registers}.
\newblock In \emph{The Twelfth International Conference on Learning Representations}, 2024.

\bibitem[Dubey et~al.(2024)Dubey, Jauhri, Pandey, Kadian, Al-Dahle, Letman, Mathur, Schelten, Yang, Fan, et~al.]{llama3}
Abhimanyu Dubey, Abhinav Jauhri, Abhinav Pandey, Abhishek Kadian, Ahmad Al-Dahle, Aiesha Letman, Akhil Mathur, Alan Schelten, Amy Yang, Angela Fan, et~al.
\newblock {The llama 3 herd of models}.
\newblock \emph{arXiv preprint arXiv:2407.21783}, 2024.

\bibitem[Hirschorn and Avidan(2024)]{poseanything}
Or Hirschorn and Shai Avidan.
\newblock A graph-based approach for category-agnostic pose estimation, 2024.

\bibitem[Hu et~al.(2022)Hu, Shen, Wallis, Allen-Zhu, Li, Wang, Wang, and Chen]{lora}
Edward~J Hu, Yelong Shen, Phillip Wallis, Zeyuan Allen-Zhu, Yuanzhi Li, Shean Wang, Lu Wang, and Weizhu Chen.
\newblock {LoRA: Low-Rank Adaptation of Large Language Models}.
\newblock In \emph{International Conference on Learning Representations}, 2022.

\bibitem[Jiang et~al.(2023)Jiang, Sablayrolles, Mensch, Bamford, Chaplot, Casas, Bressand, Lengyel, Lample, Saulnier, et~al.]{mistral}
Albert~Q Jiang, Alexandre Sablayrolles, Arthur Mensch, Chris Bamford, Devendra~Singh Chaplot, Diego de~las Casas, Florian Bressand, Gianna Lengyel, Guillaume Lample, Lucile Saulnier, et~al.
\newblock {Mistral 7B}.
\newblock \emph{arXiv preprint arXiv:2310.06825}, 2023.

\bibitem[Kirillov et~al.(2023)Kirillov, Mintun, Ravi, Mao, Rolland, Gustafson, Xiao, Whitehead, Berg, Lo, et~al.]{sam}
Alexander Kirillov, Eric Mintun, Nikhila Ravi, Hanzi Mao, Chloe Rolland, Laura Gustafson, Tete Xiao, Spencer Whitehead, Alexander~C Berg, Wan-Yen Lo, et~al.
\newblock {Segment anything}.
\newblock In \emph{Proceedings of the IEEE/CVF International Conference on Computer Vision}, pages 4015--4026, 2023.

\bibitem[Labuguen et~al.(2021)Labuguen, Matsumoto, Negrete, Nishimaru, Nishijo, Takada, Go, Inoue, and Shibata]{macaquepose}
Rollyn Labuguen, Jumpei Matsumoto, Salvador~Blanco Negrete, Hiroshi Nishimaru, Hisao Nishijo, Masahiko Takada, Yasuhiro Go, Ken-ichi Inoue, and Tomohiro Shibata.
\newblock {MacaquePose: a novel “in the wild” macaque monkey pose dataset for markerless motion capture}.
\newblock \emph{Frontiers in behavioral neuroscience}, 14:\penalty0 581154, 2021.

\bibitem[Lai et~al.(2024)Lai, Tian, Chen, Li, Yuan, Liu, and Jia]{lisa}
Xin Lai, Zhuotao Tian, Yukang Chen, Yanwei Li, Yuhui Yuan, Shu Liu, and Jiaya Jia.
\newblock {Lisa: Reasoning segmentation via large language model}.
\newblock In \emph{Proceedings of the IEEE/CVF Conference on Computer Vision and Pattern Recognition}, pages 9579--9589, 2024.

\bibitem[Li et~al.(2023)Li, Zhang, Zhang, Long, Xie, and Zhang]{gte}
Zehan Li, Xin Zhang, Yanzhao Zhang, Dingkun Long, Pengjun Xie, and Meishan Zhang.
\newblock {Towards general text embeddings with multi-stage contrastive learning}.
\newblock \emph{arXiv preprint arXiv:2308.03281}, 2023.

\bibitem[Lin et~al.(2015)Lin, Maire, Belongie, Bourdev, Girshick, Hays, Perona, Ramanan, Zitnick, and Dollár]{mscoco}
Tsung-Yi Lin, Michael Maire, Serge Belongie, Lubomir Bourdev, Ross Girshick, James Hays, Pietro Perona, Deva Ramanan, C.~Lawrence Zitnick, and Piotr Dollár.
\newblock {Microsoft COCO: Common Objects in Context}, 2015.

\bibitem[Liu et~al.(2024)Liu, Li, Wu, and Lee]{llava}
Haotian Liu, Chunyuan Li, Qingyang Wu, and Yong~Jae Lee.
\newblock {Visual instruction tuning}.
\newblock \emph{Advances in neural information processing systems}, 36, 2024.

\bibitem[Liu et~al.(2023)Liu, Zeng, Ren, Li, Zhang, Yang, Li, Yang, Su, Zhu, et~al.]{grounding_dino}
Shilong Liu, Zhaoyang Zeng, Tianhe Ren, Feng Li, Hao Zhang, Jie Yang, Chunyuan Li, Jianwei Yang, Hang Su, Jun Zhu, et~al.
\newblock {Grounding dino: Marrying dino with grounded pre-training for open-set object detection}.
\newblock \emph{arXiv preprint arXiv:2303.05499}, 2023.

\bibitem[Luo et~al.(2021)Luo, Wang, Huang, Wang, Tan, and Zhou]{luo2021rethinking}
Zhengxiong Luo, Zhicheng Wang, Yan Huang, Liang Wang, Tieniu Tan, and Erjin Zhou.
\newblock {Rethinking the heatmap regression for bottom-up human pose estimation}.
\newblock In \emph{Proceedings of the IEEE/CVF conference on computer vision and pattern recognition}, pages 13264--13273, 2021.

\bibitem[Nguyen et~al.(2024)Nguyen, Li, and Lee]{escape}
Khoi~Duc Nguyen, Chen Li, and Gim~Hee Lee.
\newblock {ESCAPE: Encoding Super-keypoints for Category-Agnostic Pose Estimation}.
\newblock In \emph{Proceedings of the IEEE/CVF Conference on Computer Vision and Pattern Recognition}, pages 23491--23500, 2024.

\bibitem[Oquab et~al.(2024)Oquab, Darcet, Moutakanni, Vo, Szafraniec, Khalidov, Fernandez, HAZIZA, Massa, El-Nouby, Assran, Ballas, Galuba, Howes, Huang, Li, Misra, Rabbat, Sharma, Synnaeve, Xu, Jegou, Mairal, Labatut, Joulin, and Bojanowski]{dinov2}
Maxime Oquab, Timoth{\'e}e Darcet, Th{\'e}o Moutakanni, Huy~V. Vo, Marc Szafraniec, Vasil Khalidov, Pierre Fernandez, Daniel HAZIZA, Francisco Massa, Alaaeldin El-Nouby, Mido Assran, Nicolas Ballas, Wojciech Galuba, Russell Howes, Po-Yao Huang, Shang-Wen Li, Ishan Misra, Michael Rabbat, Vasu Sharma, Gabriel Synnaeve, Hu Xu, Herve Jegou, Julien Mairal, Patrick Labatut, Armand Joulin, and Piotr Bojanowski.
\newblock {DINOv2: Learning Robust Visual Features without Supervision}.
\newblock \emph{Transactions on Machine Learning Research}, 2024.

\bibitem[Ranasinghe et~al.(2024)Ranasinghe, Shukla, Poursaeed, Ryoo, and Lin]{ranasinghe2024learning}
Kanchana Ranasinghe, Satya~Narayan Shukla, Omid Poursaeed, Michael~S Ryoo, and Tsung-Yu Lin.
\newblock {Learning to localize objects improves spatial reasoning in visual-llms}.
\newblock In \emph{Proceedings of the IEEE/CVF Conference on Computer Vision and Pattern Recognition}, pages 12977--12987, 2024.

\bibitem[Reddy et~al.(2018)Reddy, Vo, and Narasimhan]{carfusion}
N~Dinesh Reddy, Minh Vo, and Srinivasa~G Narasimhan.
\newblock {Carfusion: Combining point tracking and part detection for dynamic 3d reconstruction of vehicles}.
\newblock In \emph{Proceedings of the IEEE conference on computer vision and pattern recognition}, pages 1906--1915, 2018.

\bibitem[Reddy et~al.(2019)Reddy, Vo, and Narasimhan]{occlusionnet}
N~Dinesh Reddy, Minh Vo, and Srinivasa~G Narasimhan.
\newblock {Occlusion-net: 2d/3d occluded keypoint localization using graph networks}.
\newblock In \emph{Proceedings of the IEEE/CVF conference on computer vision and pattern recognition}, pages 7326--7335, 2019.

\bibitem[Ren et~al.(2024)Ren, Gao, Sun, Qi, Wang, and Liao]{dynamic_support_info}
Pengfei Ren, Yuanyuan Gao, Haifeng Sun, Qi Qi, Jingyu Wang, and Jianxin Liao.
\newblock {Dynamic Support Information Mining for Category-Agnostic Pose Estimation}.
\newblock In \emph{Proceedings of the IEEE/CVF Conference on Computer Vision and Pattern Recognition}, pages 1921--1930, 2024.

\bibitem[Rusanovsky et~al.(2024)Rusanovsky, Hirschorn, and Avidan]{capex}
Matan Rusanovsky, Or Hirschorn, and Shai Avidan.
\newblock {CapeX: Category-Agnostic Pose Estimation from Textual Point Explanation}.
\newblock \emph{arXiv preprint arXiv:2406.00384}, 2024.

\bibitem[Ryali et~al.(2023)Ryali, Hu, Bolya, Wei, Fan, Huang, Aggarwal, Chowdhury, Poursaeed, Hoffman, et~al.]{hiera}
Chaitanya Ryali, Yuan-Ting Hu, Daniel Bolya, Chen Wei, Haoqi Fan, Po-Yao Huang, Vaibhav Aggarwal, Arkabandhu Chowdhury, Omid Poursaeed, Judy Hoffman, et~al.
\newblock {Hiera: A hierarchical vision transformer without the bells-and-whistles}.
\newblock In \emph{International Conference on Machine Learning}, pages 29441--29454. PMLR, 2023.

\bibitem[Shi et~al.(2023)Shi, Huang, Ma, Hu, and Cao]{capeformer}
Min Shi, Zihao Huang, Xianzheng Ma, Xiaowei Hu, and Zhiguo Cao.
\newblock {Matching is not enough: A two-stage framework for category-agnostic pose estimation}.
\newblock In \emph{Proceedings of the IEEE/CVF Conference on Computer Vision and Pattern Recognition}, pages 7308--7317, 2023.

\bibitem[Song and Bahri(2025)]{song2025decoding}
Xingyou Song and Dara Bahri.
\newblock Decoding-based regression.
\newblock \emph{arXiv preprint arXiv:2501.19383}, 2025.

\bibitem[Sun et~al.(2019)Sun, Xiao, Liu, and Wang]{hrnet}
Ke Sun, Bin Xiao, Dong Liu, and Jingdong Wang.
\newblock {Deep high-resolution representation learning for human pose estimation}.
\newblock In \emph{Proceedings of the IEEE/CVF conference on computer vision and pattern recognition}, pages 5693--5703, 2019.

\bibitem[Wang et~al.(2024{\natexlab{a}})Wang, Xuan, and Zhang]{locllm}
Dongkai Wang, Shiyu Xuan, and Shiliang Zhang.
\newblock {LocLLM: Exploiting Generalizable Human Keypoint Localization via Large Language Model}.
\newblock In \emph{Proceedings of the IEEE/CVF Conference on Computer Vision and Pattern Recognition}, pages 614--623, 2024{\natexlab{a}}.

\bibitem[Wang et~al.(2024{\natexlab{b}})Wang, Chen, Chen, Wu, Zhu, Zeng, Luo, Lu, Zhou, Qiao, et~al.]{visionllm}
Wenhai Wang, Zhe Chen, Xiaokang Chen, Jiannan Wu, Xizhou Zhu, Gang Zeng, Ping Luo, Tong Lu, Jie Zhou, Yu Qiao, et~al.
\newblock {Visionllm: Large language model is also an open-ended decoder for vision-centric tasks}.
\newblock \emph{Advances in Neural Information Processing Systems}, 36, 2024{\natexlab{b}}.

\bibitem[Wei et~al.(2023)Wei, Zhang, Zhang, Zhang, and Chu]{lenna}
Fei Wei, Xinyu Zhang, Ailing Zhang, Bo Zhang, and Xiangxiang Chu.
\newblock {Lenna: Language enhanced reasoning detection assistant}.
\newblock \emph{arXiv preprint arXiv:2312.02433}, 2023.

\bibitem[Wei et~al.(2022)Wei, Wang, Schuurmans, Bosma, Xia, Chi, Le, Zhou, et~al.]{cot}
Jason Wei, Xuezhi Wang, Dale Schuurmans, Maarten Bosma, Fei Xia, Ed Chi, Quoc~V Le, Denny Zhou, et~al.
\newblock {Chain-of-thought prompting elicits reasoning in large language models}.
\newblock \emph{Advances in neural information processing systems}, 35:\penalty0 24824--24837, 2022.

\bibitem[Wu et~al.(2024{\natexlab{a}})Wu, Li, Yu, Wang, Chen, Gu, Yao, Shang, and McAuley]{commit}
Junda Wu, Xintong Li, Tong Yu, Yu Wang, Xiang Chen, Jiuxiang Gu, Lina Yao, Jingbo Shang, and Julian McAuley.
\newblock {Commit: Coordinated instruction tuning for multimodal large language models}.
\newblock \emph{arXiv preprint arXiv:2407.20454}, 2024{\natexlab{a}}.

\bibitem[Wu et~al.(2024{\natexlab{b}})Wu, Zhong, Xing, Lai, Liu, Chen, Wang, Zhu, Lu, Lu, Luo, Qiao, and Dai]{visionllmv2}
Jiannan Wu, Muyan Zhong, Sen Xing, Zeqiang Lai, Zhaoyang Liu, Zhe Chen, Wenhai Wang, Xizhou Zhu, Lewei Lu, Tong Lu, Ping Luo, Yu Qiao, and Jifeng Dai.
\newblock {Vision{LLM} v2: An End-to-End Generalist Multimodal Large Language Model for Hundreds of Vision-Language Tasks}.
\newblock In \emph{The Thirty-eighth Annual Conference on Neural Information Processing Systems}, 2024{\natexlab{b}}.

\bibitem[Xiao et~al.(2018)Xiao, Wu, and Wei]{simplebaseline}
Bin Xiao, Haiping Wu, and Yichen Wei.
\newblock {Simple baselines for human pose estimation and tracking}.
\newblock In \emph{Proceedings of the European conference on computer vision (ECCV)}, pages 466--481, 2018.

\bibitem[Xu et~al.(2025)Xu, Jin, Li, Song, Sun, and Yuan]{llava-cot}
Guowei Xu, Peng Jin, Hao Li, Yibing Song, Lichao Sun, and Li Yuan.
\newblock {LLaVA-CoT: Let Vision Language Models Reason Step-by-Step}, 2025.

\bibitem[Xu et~al.(2022{\natexlab{a}})Xu, Jin, Zeng, Liu, Qian, Ouyang, Luo, and Wang]{pomnet}
Lumin Xu, Sheng Jin, Wang Zeng, Wentao Liu, Chen Qian, Wanli Ouyang, Ping Luo, and Xiaogang Wang.
\newblock {Pose for everything: Towards category-agnostic pose estimation}.
\newblock In \emph{European conference on computer vision}, pages 398--416. Springer, 2022{\natexlab{a}}.

\bibitem[Xu et~al.(2022{\natexlab{b}})Xu, Zhang, Zhang, and Tao]{vitpose}
Yufei Xu, Jing Zhang, Qiming Zhang, and Dacheng Tao.
\newblock {Vitpose: Simple vision transformer baselines for human pose estimation}.
\newblock \emph{Advances in Neural Information Processing Systems}, 35:\penalty0 38571--38584, 2022{\natexlab{b}}.

\bibitem[Yu et~al.(2021)Yu, Xu, Zhang, Zhao, Guan, and Tao]{ap10k}
Hang Yu, Yufei Xu, Jing Zhang, Wei Zhao, Ziyu Guan, and Dacheng Tao.
\newblock {AP-10K: A Benchmark for Animal Pose Estimation in the Wild}.
\newblock In \emph{Thirty-fifth Conference on Neural Information Processing Systems Datasets and Benchmarks Track (Round 2)}, 2021.

\bibitem[Zheng et~al.(2023)Zheng, Chiang, Sheng, Zhuang, Wu, Zhuang, Lin, Li, Li, Xing, et~al.]{vicuna}
Lianmin Zheng, Wei-Lin Chiang, Ying Sheng, Siyuan Zhuang, Zhanghao Wu, Yonghao Zhuang, Zi Lin, Zhuohan Li, Dacheng Li, Eric Xing, et~al.
\newblock {Judging llm-as-a-judge with mt-bench and chatbot arena}.
\newblock \emph{Advances in Neural Information Processing Systems}, 36:\penalty0 46595--46623, 2023.

\bibitem[Zhu et~al.(2024)Zhu, Chen, Shen, Li, and Elhoseiny]{minigpt}
Deyao Zhu, Jun Chen, Xiaoqian Shen, Xiang Li, and Mohamed Elhoseiny.
\newblock {Mini{GPT}-4: Enhancing Vision-Language Understanding with Advanced Large Language Models}.
\newblock In \emph{The Twelfth International Conference on Learning Representations}, 2024.

\end{thebibliography}
}
\clearpage
\setcounter{section}{0}
\maketitlesupplementary
\renewcommand{\thesection}{\Alph{section}}

\section{Keypoint Descriptions}
\label{appdx:def_name_desc}

We create the names and descriptions of keypoints for all 100 categories. The names can be divided into two types: one that has its own unique name, e.g., \texttt{left shoulder}, \texttt{right eye}, and the other that does not have its own name. the latter is difficult to define due to the densely distributed position. We concentrate on designating the latter and determine the names using their relative positions in each category; for example, ``upper", ``central", ``lower''. The descriptions are represented with the keypoint position in the category and its relation with other keypoints; e.g., in the \textit{animal body}, the description of \texttt{left front paw} is defined as ``The left front paw is the lower end of the left forelimb, used for movement and manipulation of objects. It is positioned below the left elbow and connected with the left elbow''. A detailed example can be found in Table~\ref{tab:appdx/def_name_desc/instruction_example}.

\section{Exploring other Design Choices}
\label{appdx:design_choices}

\subsection{Instruction}
\label{appdx:design_choices/instruction}

\begin{table}[htbp!]
  \centering
  \resizebox{\linewidth}{!}{
  \begin{tabular}{ccccc}
    \toprule
    \textbf{w/ description} & \textbf{w/ keypoint list} & \textbf{PCK@0.05} & \textbf{PCK@0.2} & \textbf{mPCK} \\
    \midrule
    \redx & \redx & 72.60 & 96.22 & 89.86 \\
    \greencheck & \redx & \cellcolor{mylightblue} \textbf{78.43} & \cellcolor{mylightblue} \textbf{96.98} & \cellcolor{mylightblue} \textbf{91.98} \\
    \greencheck & \greencheck & 77.36 & 95.80 & 90.97 \\
    \bottomrule
  \end{tabular}
}
\vspace{-0.3cm}
\caption{Effect of additional info for keypoints in training. \colorbox{mylightblue}{Default config}.}
\vspace{-0.5cm}
\label{tab:appdx/design_choices/instruction/without_description}
\end{table}

\begin{table}[htbp!]
  \centering
  \resizebox{\linewidth}{!}{
  \begin{tabular}{ccccc}
    \toprule
    \textbf{Diverse questions} & \textbf{Add conversation outline} & \textbf{PCK@0.05} & \textbf{PCK@0.2} & \textbf{mPCK} \\
    \midrule
    \redx & \redx & \cellcolor{mylightblue} \textbf{78.43}& \cellcolor{mylightblue} \textbf{96.98} & \cellcolor{mylightblue} \textbf{91.98} \\
    \greencheck & \redx & 74.24 & 96.56 & 90.56 \\
    \redx & \greencheck & 75.08 & 96.27 & 90.63 \\
    \greencheck & \greencheck & 68.24 & 95.93 & 88.52 \\
    \bottomrule
  \end{tabular}
}
\vspace{-0.3cm}
\caption{Effect of adding a conversation outline and diversifying question expressions. \colorbox{mylightblue}{Default config}.}
\vspace{-0.5cm}
\label{tab:appdx/design_choices/instruction/random_question_system_info}
\end{table}

\paragraph{Instruction variations}
As mentioned in Sec~\ref{sec:method/instruction}, we include not only the names but also descriptions of the keypoints in the instructions to help the model better to reason the location of keypoints. We examine how the description affects model performance by training the model without descriptions. The result in Table~\ref{tab:appdx/design_choices/instruction/without_description} shows that without descriptions, the accuracy decreases over 2\%p in mPCK, suggesting that the keypoint description plays a significant role in enhancing to find the exact position. We experiment another scenario to include all keypoint names for each category in the instruction as ``Keypoint List". As shown in Table~\ref{tab:appdx/design_choices/instruction/without_description}, unlike keypoint descriptions, the list of keypoint names is not helpful for improving the model, rather reducing its performance.
Next, we explore whether two optional conditions affect the performance or not: one is encompassing a conversation outline~\cite{llava} and the other is to diversify the question expression in instruction. The outline slightly modified from the prior work~\cite{llava} seems not to influence to solve the problem that predicts coordinates, and the random question does not have any positive effect on the performance, actually leading to a decrease in the model's performance(Table~\ref{tab:appdx/design_choices/instruction/random_question_system_info}).

\begin{table}[htbp!]
    \centering
    \resizebox{\linewidth}{!}{
        \begin{tabular}{ccccccc}
            \toprule
            \textbf{Multi-round} & \textbf{PCK\@0.05} & \textbf{PCK\@0.10} & \textbf{PCK\@0.15} & \textbf{PCK\@0.20} & \textbf{PCK\@0.25} & \textbf{mPCK} \\
            \midrule
            $k$ = 1 & 78.29 & 91.55 & 95.19 & 96.89 & 97.88 & 91.96 \\
            \midrule
            $k$ = 2 & 72.82 & 88.06 & 92.79 & 95.30 & 96.56 & 89.11 \\
            \midrule
            \rowcolor{mylightblue} $k$ = 4 & \textbf{78.43} & \textbf{91.34} & \textbf{95.26} & \textbf{96.98} & \textbf{97.90} & \textbf{91.98} \\
            \midrule
            $k$ = 6 & 74.33 & 89.82 & 94.17 & 96.36 & 97.46 & 90.43 \\
            \midrule
            $k$ = 8 & 75.28 & 89.89 & 93.99 & 96.16 & 97.41 & 90.55 \\
            \bottomrule
            \end{tabular}
    }
    \vspace{-0.3cm}
    \caption{Ablation in multi-round $k$. \colorbox{mylightblue}{Default config}.}
    \label{tab:appdx/design_choices/instruction/multi-round-k}
    \vspace{-0.5cm}
\end{table}

\paragraph{Choice of round $k$}
We investigate the optimal number of rounds $k$ in the conversation. Table~\ref{tab:appdx/design_choices/instruction/multi-round-k} shows that under the same training conditions, the highest performance was observed when $k$ is set to 4. No explicit tendency was found as $k$ changed.

\begin{table}[htbp]
  \centering
  \resizebox{\linewidth}{!}{
    \begin{tabular}{lcccc}
    \toprule
    \textbf{LLM} & \textbf{Step-by-step instruction} & \textbf{PCK@0.05} & \textbf{PCK@0.2} & \textbf{mPCK} \\
    \midrule
    \multirow{2}{*}{Llama3.1-8B~\cite{llama3}} & \redx & \cellcolor{mylightblue} \textbf{78.43} & \cellcolor{mylightblue} \textbf{96.98} & \cellcolor{mylightblue} \textbf{91.98} \\
    & \greencheck & 76.06 & 96.48 & 91.11 \\
    \midrule
    \multirow{2}{*}{Llama3.2-1B~\cite{Llama3.2}} & \redx & 76.46 & 96.41 & 91.20 \\
    & \greencheck & 76.65 & 96.75 & 91.49 \\
    \bottomrule
  \end{tabular}
}
\caption{Performance comparison with \textit{step-by-step instruction} across different LLMs. \colorbox{mylightblue}{Default config}.}
\vspace{-0.5cm}
\label{tab:appdx/design_choices/instruction/step_by_step_instruction}
\end{table}

\paragraph{Different style of instruction}
We take another structure of instruction question-answering in a step-by-step manner, so-called \textit{step-by-step instruction}(Figure~\ref{fig:appdx/design_choices/instruction/step-by-step-instruction}). Specifically, Rather than providing instruction as Figure~\ref{fig:model_arch}, we question what the object is and then inquire the coordinates of keypoints. We expect this approach would help the model better understand the input. Interestingly, the effect of this mechanism varies depending on the LLM, as in Table~\ref{tab:appdx/design_choices/instruction/step_by_step_instruction}. It appears that different LLMs require different approaches to better understand the instruction.

\subsection{Architecture}
\label{appdx:design_choices/architecture}

\begin{table}[htbp!]
    \centering
    \resizebox{\linewidth}{!}{
        \begin{tabular}{ccccccc}
            \toprule
            \textbf{Visual Encoder} & \textbf{PCK\@0.05} & \textbf{PCK\@0.10} & \textbf{PCK\@0.15} & \textbf{PCK\@0.20} & \textbf{PCK\@0.25} & \textbf{mPCK} \\
            \midrule
            DINO-v2-reg~\cite{dinov2-reg} & 62.52 & 86.00 & 92.83 & 95.83 & 97.34 & 86.90 \\
            \midrule
            Hiera~\cite{hiera} & 56.13 & 83.31 & 91.99 & 95.67 & 97.35 & 84.89 \\
            \midrule
            \rowcolor{mylightblue} DINO-v2~\cite{dinov2} & \textbf{78.43} & \textbf{91.34} & \textbf{95.26} & \textbf{96.98} & \textbf{97.90} & \textbf{91.98} \\
            \bottomrule
            \end{tabular}
    }
    \vspace{-0.3cm}
    \caption{Ablation in visual encoders. \colorbox{mylightblue}{Default config}.}
    \label{tab:appdx/design_choices/architecture/different_visual_encoder}
\end{table}

\begin{table}[htbp!]
    \centering
    \resizebox{\linewidth}{!}{
        \begin{tabular}{ccccccc}
            \toprule
            \textbf{Fine-tuning method} & \textbf{PCK\@0.05} & \textbf{PCK\@0.10} & \textbf{PCK\@0.15} & \textbf{PCK\@0.20} & \textbf{PCK\@0.25} & \textbf{mPCK} \\
            \midrule
            None (Frozen) & 69.69 & 88.16 & 92.62 & 95.07 & 96.41 & 88.39 \\
            \midrule
            \rowcolor{mylightblue} LoRA~\cite{lora} & \textbf{78.43} & \textbf{91.34} & \textbf{95.26} & \textbf{96.98} & \textbf{97.90} & \textbf{91.98} \\
            \midrule
            Full parameters & 6.93 & 23.72 & 42.41 & 55.31 & 64.56 & 38.59 \\
            \bottomrule
            \end{tabular}
    }
    \vspace{-0.3cm}
    \caption{Ablation in fine-tuning methods. \colorbox{mylightblue}{Default config}.}
    \label{tab:appdx/design_choices/architecture/different_visual_encoder_tuning}
    \vspace{-0.5cm}
\end{table}

\paragraph{Choice of visual encoder}
We conduct an ablation study for the visual encoder in CapeLLM. We choose three popular visual encoders: DINO-v2~\cite{dinov2}, Hiera~\cite{hiera}, DINO-v2-reg~\cite{dinov2-reg}, which are pre-trained on same dataset. Table~\ref{tab:appdx/design_choices/architecture/different_visual_encoder} shows that using DINO-v2~\cite{dinov2} yields the highest performance. The known issue in DINO-v2, artifacts in the feature maps~\cite{dinov2-reg}), seems to have little  impact on performance in the CAPE task. A noteworthy point is the number of image tokens. Although Hiera~\cite{hiera} has 20\% less image tokens than the other two encoders, the performance gap is just about 1\%p, implying that retaining a larger number of image tokens does not necessarily have something to do with performance. Then, we examine three types of fine-tuning methods: full fine-tuning, fine-tuning with LoRA~\cite{lora}, and freezing. In constrat with the traditional MLLMs~\cite{llava, minigpt, visionllm}, visual encoder with LoRA was more advantageous than the other two options as~\cite{locllm}(Table~\ref{tab:appdx/design_choices/architecture/different_visual_encoder_tuning}). Notably, the full fine-tuning approach, where all parameters are learnable, drastically deteriorate the performance. This fact seems to imply that  when using relatively small datasets, leaving all parameters trainable may lead to overfitting, thus resulting in severe degradation in performance.

\begin{table}[htbp!]
    \centering
    \resizebox{\linewidth}{!}{
        \begin{tabular}{ccccccc}
            \toprule
            \textbf{LLM} & \textbf{PCK\@0.05} & \textbf{PCK\@0.10} & \textbf{PCK\@0.15} & \textbf{PCK\@0.20} & \textbf{PCK\@0.25} & \textbf{mPCK} \\
            \midrule
            Llama3.2-1B~\cite{Llama3.2} & 76.46 & 91.05 & 94.69 & 96.41 & 97.40 & 91.20 \\
            \midrule
            Vicuna-7B-v1.5~\cite{vicuna} & 62.15 & 84.40 & 91.51 & 94.79 & 96.33 & 85.84 \\
            \midrule
            Mistral-7B-v0.3~\cite{mistral} & 77.63 & 91.32 & 94.90 & 96.46 & 97.54 & 91.57 \\
            \midrule
            \rowcolor{mylightblue} Llama3.1-8B~\cite{llama3} & \textbf{78.43} & \textbf{91.34} & \textbf{95.26} & \textbf{96.98} & \textbf{97.90} & \textbf{91.98} \\
            \bottomrule
            \end{tabular}
    }
    \vspace{-0.3cm}
    \caption{Ablation in LLM. \colorbox{mylightblue}{Default config}.}
    \label{tab:appdx/design_choices/architecture/different_llms}
    \vspace{-0.5cm}
\end{table}

\paragraph{Choice of LLM}
To analyze the performance variations coming from different LLMs, we select four most recent and popular language models: Vicuna-7B~\cite{vicuna}, Mistral-7B~\cite{mistral}, Llama3.1-8B~\cite{llama3}, and Llama3.2-1B~\cite{Llama3.2}. We find that the overall accuracy gets improved as the size of the LLM increases( Table~\ref{tab:appdx/design_choices/architecture/different_llms}). Exceptionally, Llama3.2-1B~\cite{Llama3.2} exhibits an overwhelming result surpassing that of a 7B-sized LLM, Vicuna-7B-v1.5, which appears to be the effect of effectively transferring the knowledge of a larger model through distillation training methods~\cite{Llama3.2}. A larger vocabulary size seems to play a essential role to positively influence the integration of visual information and language.

\begin{table}[htbp!]
    \centering
    \resizebox{\linewidth}{!}{
        \begin{tabular}{ccccc}
            \toprule
            \textbf{Instruction} & \textbf{Output format} & \textbf{PCK@0.05} & \textbf{PCK@0.2} & \textbf{mPCK} \\
            \midrule
            \multirow{2}{*}{Base instruction} & text & \cellcolor{mylightblue} \textbf{78.43} & \cellcolor{mylightblue} \textbf{96.98} & \cellcolor{mylightblue} \textbf{91.98} \\
            & special token & 76.06 & 96.48 & 91.11 \\
            \midrule
            \multirow{2}{*}{Step-by-step instruction} & text & 76.46 & 96.41 & 91.20 \\
            & special token & 76.65 & 96.75& 91.49 \\
            \bottomrule
            \end{tabular}
    }
    \vspace{-0.3cm}
    \caption{Comparison with token output format. \colorbox{mylightblue}{Default config}.}
    \label{tab:appdx/design_choices/architecture/different_output_format}
\end{table}

\begin{figure*}[htbp]
  \centering
   \includegraphics[width=0.9\linewidth, height=0.25\linewidth]{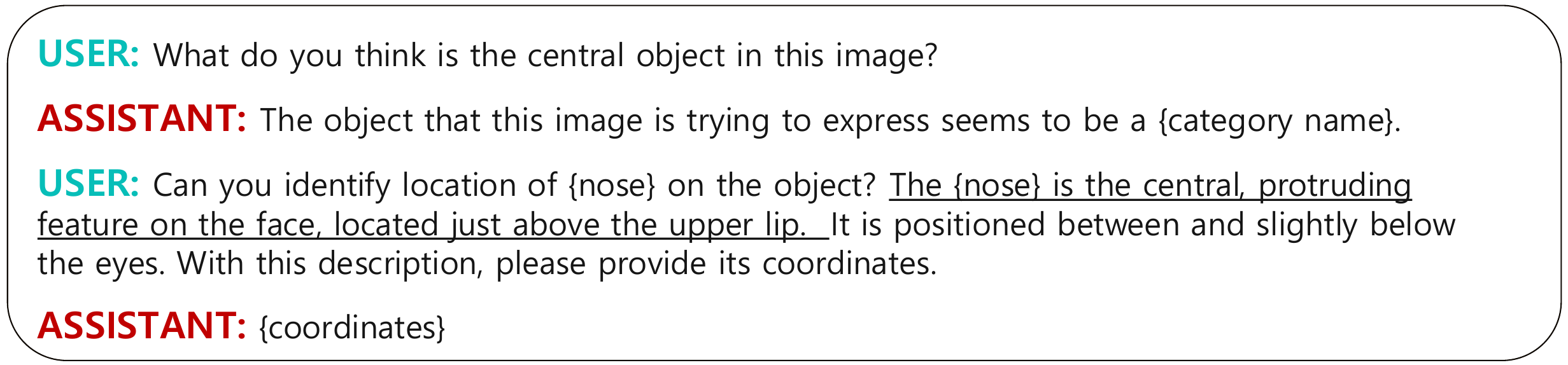}
   \caption{Step-by-step instruction. The nose is in the example above, which can be replaced with whatever you want to find out. The underlined is the description of nose, which can also be replaced according to the keypoint.}
   \label{fig:appdx/design_choices/instruction/step-by-step-instruction}
   \vspace{-0.5cm}
\end{figure*}

\paragraph{Token output format}
We explore a method that utilizes token embeddings \texttt{<KEYPOINT>} instead of text-based outputs. To introduce this method to our pipeline, some modifications in instruction should be made: the coordinates are replaced with special token \texttt{<KEYPOINT>} as answers, accordingly the vocabulary size increases, and input embeddings are turned into the trainables. The tokens are turned into the output embeddings from the LLM and are fed into a task-specific decoder. Typically, while a grounding-based pre-trained decoder is used in some tasks~\cite{lenna, lisa, visionllmv2}, no suitable decoders exist for CAPE. So, we create a simple decoder that transforms the embeddings into the coordinates and train it from scratch. We validate this method on both default instruction(as Figure~\ref{fig:model_arch}) and step-by-step one(Figure~\ref{fig:appdx/design_choices/instruction/step-by-step-instruction}). Despite the lack of pre-training, the method using \texttt{<KEYPOINT>} outputs comparable result to models with default architecture(Table~\ref{tab:appdx/design_choices/architecture/different_output_format}).

\begin{table}[t]
    \centering
    \resizebox{0.9\linewidth}{!}{
    \begin{tabular}{lcccccc}
    \toprule
    \textbf{Pre-training method} & \textbf{PCK@0.05} & \textbf{PCK@0.2} & \textbf{mPCK} \\
    \midrule
    \rowcolor{mylightblue} w/o pre-training & \textbf{78.43} & \textbf{96.98} & \textbf{91.98} \\
    Direct QA & 78.98 & 96.60 & 91.96 \\
    Step-by-step QA & 78.05 & 96.23 & 91.40 \\
    \bottomrule
    \end{tabular}
    }
\caption{Comparison in pre-training methods. \colorbox{mylightblue}{Default config}.}
\vspace{-0.5cm}
\label{tab:appdx/training_strategies}
\end{table}

\section{Pre-Training Strategy}
\label{appdx:other_training_strategy}
\begin{figure*}[htbp]
  \centering
   \includegraphics[width=\linewidth, height=0.15\linewidth]{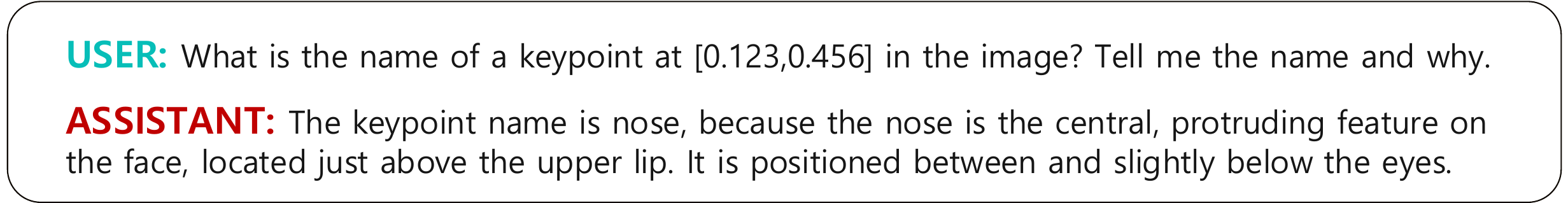}
   \caption{Instruction of direct QA for pre-training.}
   \label{fig:appdx/other_training_strategy/direct_instruction}
   \vspace{-0.5cm}
\end{figure*}

\begin{figure*}[htbp]
  \centering
   \includegraphics[width=\linewidth, height=0.5\linewidth]{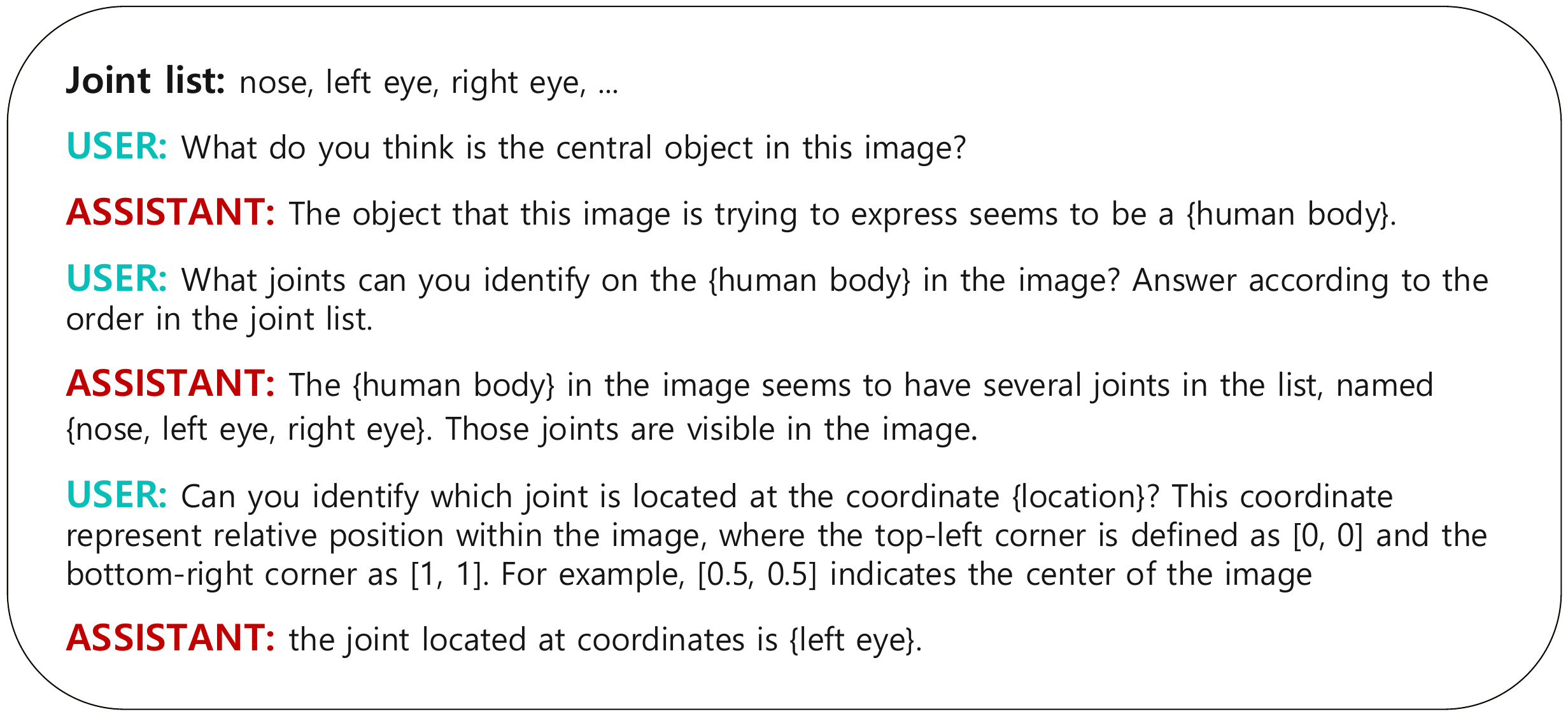}
   \caption{Instruction of step-by-step QA for pre-training.}
   \label{fig:appdx/other_training_strategy/step_by_step_instruction}
\end{figure*}

We attempt two types of pre-training process: \textit{direct QA} and \textit{step-by-step QA}. The \textit{direct QA} has an instruction that it is in the form of asking and answering the name of the keypoint corresponding to the coordinates, as in Figure~\ref{fig:appdx/other_training_strategy/direct_instruction}. On the other hand, step-by-step QA in Figure~\ref{fig:appdx/other_training_strategy/step_by_step_instruction} has an instruction that is in the form of asking about the category, inquiring the existence of the keypoint in the image, and then inducing the selection of the keypoint corresponding to the coordinates. Referring to the related works~\cite{ranasinghe2024learning, visionllm, visionllmv2}, all layers except for projection layer are frozen in this stage. As a consequence, there is no positive effect on the performance gain, as shown in Table~\ref{tab:appdx/training_strategies}. In light of the use of large-scale pre-training data in the previous methods~\cite{ranasinghe2024learning, visionllm, locllm, visionllmv2}, we conjecture that the limited number of images in each category might result in this outcome.

\vspace{-0.1cm}

\begin{table*}[htbp]
\centering
\begin{tabular}{lp{0.7\textwidth}}
\toprule
\textbf{Keypoint} & \textbf{Description} \\
\midrule
Left eye & The left eye is one of the two visual organs located on the face. It is positioned slightly to the left of the nose and just below the brow ridge, visible from the front. \\
\midrule
Right eye & The right eye is the visual organ located on the right side of the face. It is situated to the right of the nose and directly opposite the left eye. \\
\midrule
Nose & The nose is the central, protruding feature on the face, located just above the upper lip. It is positioned between and slightly below the eyes \\
\midrule
Neck & The neck is the part of the body connecting the head to the torso that refers to the area from the shoulders to the hip joints. It is located below the head, near the junction where the shoulders meet the body. \\
\midrule
Root of tail & The root of the tail is at the base of the spine, where the tail begins. It is located near the lower back, above the hips. \\
\midrule
Left shoulder & The left shoulder is the joint connecting the left arm to the torso. It is situated to the left of the neck and above the left elbow. \\
\midrule
Left elbow & The left elbow is the joint in the middle of the left arm, connecting the upper arm to the forearm. It is located between the left shoulder and the left front paw and connectd with them. \\
\midrule
Left front paw & The left front paw is the lower end of the left forelimb, used for movement and manipulation of objects. It is positioned below the left elbow and connected with the left elbow. \\
\midrule
Right shoulder & The right shoulder is the joint connecting the right arm to the torso. It is located to the right of the neck and above the right elbow. \\
\midrule
Right elbow & The right elbow is the joint in the middle of the right arm, connecting the upper arm to the forearm. It is situated between the right shoulder and the right front paw and connectd with them. \\
\midrule
Right front paw & The right front paw is the lower end of the right forelimb, used for movement and manipulation of objects. It is located below the right elbow and connectd with the right elbow. \\
\midrule
Left hip & The left hip is the joint connecting the left leg to the torso. It is positioned below the root of the tail and above the left knee. \\
\midrule
Left knee & The left knee is the joint in the middle of the left leg, connecting the upper leg to the lower leg. It is located between the left hip and the left back paw and connectd with them.. \\
\midrule
Left back paw & The left back paw is the lower end of the left hind limb, used for movement and support. It is situated below the left knee. \\
\midrule
Right hip & The right hip is the joint connecting the right leg to the torso. It is positioned below the root of the tail and above the right knee. \\
\midrule
Right knee & The right knee is the joint in the middle of the right leg, connecting the upper leg to the lower leg. It is located between the right hip and the right back paw and connectd with them. \\
\midrule
Right back paw & The right back paw is the lower end of the right hind limb, used for movement and support. It is situated below the right knee. \\
\bottomrule
\end{tabular}
\caption{An example of descriptions: \textit{animal body}}
\label{tab:appdx/def_name_desc/instruction_example}
\end{table*}

\vspace{-1cm}

\end{document}